 \documentclass[preprint,review,12pt]{elsarticle}

\usepackage{amsmath,amsthm,amssymb,amsfonts,bbm,mathrsfs,algorithmic,algorithm}
\usepackage{array,multirow,multicol,color,xcolor,setspace,caption,subcaption,url,booktabs}
\usepackage{lineno}
\newtheorem{theorem}{Theorem}
\newtheorem{lemma}{Lemma}

\newtheorem*{proof*}{Proof}
\newtheorem*{remark*}{Remark}

\journal{Signal Processing}

\begin{document}

\begin{frontmatter}



\title{Provable Tensor Ring Completion}
\author[label]{Huyan Huang}
\ead{huyanhuang@gmail.com}
\author[label]{Jiani Liu}
\ead{jianiliu@std.uestc.edu.cn}
\author[label]{Yipeng Liu\corref{cor1}\fnref{label1}}
\ead{yipengliu@uestc.edu.cn}
\cortext[cor1]{Corresponding author.}
\fntext[label1]{This research is supported by National Natural Science Foundation of China (NSFC, No. 61602091, No. 61571102).}
\author[label]{Ce Zhu}
\ead{eczhu@uestc.edu.cn}
\address[label]{School of Information and Communication Engineering, University of Electronic Science and Technology of China (UESTC), Chengdu, 611731, China.}


%\address{}

\begin{abstract}
Tensor completion recovers a multi-dimensional array from a limited number of measurements. Using the recently proposed tensor ring (TR) decomposition, in this paper we show that a $d$-order tensor of size $n\times \dotsm \times n$ and TR rank $\left[r,\dotsc,r\right]$ can be exactly recovered with high probability by solving a convex optimization program, given $O\left(n^{\lceil d/2 \rceil}r^2\ln^7\left(n^{\lceil d/2 \rceil}\right)\right)$ samples. The proposed TR incoherence condition under which the result holds is similar to the matrix incoherence condition. The experiments on synthetic data verify the recovery guarantee for TR completion. Moreover, the experiments on real-world data show that our method improves the recovery performance compared with the state-of-the-art methods.
\end{abstract}


\end{frontmatter}


\section{Introduction}
\label{section-introduction}

Tensors are natural representations for multi-dimensional data \cite{kolda2009tensor, cichocki2015tensor, cichocki2016tensor}. In the mathematical discipline of multi-linear algebra, one fundamental problem is how to express a tensor as a sequence of elementary operations acting on other simpler tensors (often interpretable). Any scheme that achieves this goal is called tensor decomposition. Tensor decomposition can capture the interactions between different modes of multi-dimensional data, thus it provides a reasonable and advantageous mathematical framework for formulating and solving problems in a range of applications, such as signal processing \cite{cichocki2015tensor}, machine learning \cite{sidiropoulos2017tensor}, remote sensing \cite{signoretto2011tensor, he2019total}, computer vision \cite{liu2013tensor}, etc.
 
Tensor completion aims to interpolate the missing entries from partially observed tensors \cite{gandy2011tensor}. One major theoretical issue in this field concerns the sampling condition for tensor completion which depends on the algebraic structure of tensor decomposition. For instance, based on CANDECOMP/PARAFAC (CP) decomposition which represents a tensor as a sum of rank-$1$ tensors \cite{kolda2009tensor}, the method proposed in \cite{mu2014square} can recover a $d$-order tensor of size $n\times \dotsm \times n$ and CP rank $r$, as long as the number of Gaussian measurements is on the order of $n^{\lceil d/2 \rceil}r$. However, determining CP rank is NP-hard \cite{hillar2013most} and a low CP rank approximation may involve numerical problems \cite{cichocki2015tensor}. Based on Tucker (TK) decomposition which factorizes a tensor into a set of matrices and one small core tensor \cite{kolda2009tensor}, \cite{tomioka2011statistical} claims that it requires $O\left(n^{d-1}r\right)$ Gaussian measurements to recover a $d$-order tensor of size $n\times \dotsm \times n$ and TK rank $\left[r,\dotsc,r\right]$. Moreover, the number of samples required to recover a tensor by TK decomposition is further reduced to $O\left(n^{\lceil d/2 \rceil}r^{\lfloor d/2 \rfloor}\right)$ via a balance unfolding scheme \cite{mu2014square}. Based on another widely used factorization namely tensor singular value decomposition (t-SVD) \cite{kilmer2013third}, the authors of \cite{zhang2017exact} show that a $d$-order tensor of size $n\times \dotsm \times n$ and tubal rank $r$ can be exactly recovered, so long as $O\left(n^{d-1}r\ln\left(n^{d-1}\right)\right)$ entries are sampled under random sampling. Methods based on tensor train (TT) decomposition and hierarchical Tucker (HT) decomposition under random sampling can refer to \cite{bengua2017efficient} and \cite{liu2019image}.

The recently proposed tensor ring (TR) decomposition represents a high-order tensor as several cyclically contracted $3$-order tensors \cite{zhao2017learning, ye2018tensor}, which is a linear combination of TT. The first TR decomposition based method is proposed in \cite{wang2017efficient}, in which the completion model is formulated as a data fitting problem for the given partial observations. The algorithm optimizes each latent TR factor alternately. However, this method suffers from expensive time cost and overfitting problem when a smaller number of samples are available, and its performance highly rests on the choice of TR rank. Subsequently, a gradient descent method for TR completion is proposed in \cite{yuan2018higher}, where all TR factors are simultaneously optimized in one iteration. This method reduces computational cost but still requires a pre-defined TR rank. By exploiting the low rank structure of the TR latent space, a nuclear norm regularization model is propounded to alleviate the burden of TR rank selection \cite{yuan2019tensor}, which greatly reduces the computational cost. In \cite{yu2019tensor}, a TR nuclear norm minimization model with tensor circular unfolding scheme is proposed for tensor completion. This method does not require a pre-defined TR rank and achieves better performance than previous TR decomposition based methods.

However, existing TR decomposition based completion methods do not have theoretical guarantee. In this paper, by leveraging the McDiarmid inequality, we prove that most tensors satisfy TR incoherence property if we constrain the mode-$2$ fibers of TR factors to be incoherent. Using an $i$-shifting $l$-matricization scheme \cite{huang2019provable}, we propose a TR nuclear norm minimization model for tensor completion with random sampling. We show that the proposed method can recover a $d$-order tensor of size $n\times \dotsm \times n$ and TR rank $\left[r,\dotsc,r\right]$ with high probability under the TR incoherence condition, given $O\left(n^{\lceil d/2 \rceil}r^2\ln^7\left(n^{\lceil d/2 \rceil}\right)\right)$ samples. The proposed theory and the effectiveness of the algorithm are confirmed by experiments on synthetic data and real-world data.

The rest of this paper is organized as follows. Section \ref{section-notation} provides basic notations and preliminaries of TR decomposition. In Section \ref{section-sampling}, we propose a weighted sum of nuclear norm model for tensor completion with recovery guarantee under random sampling. The corresponding proof is in Section \ref{section-proof}. Section \ref{section-experiment} exhibits the results of numerical experiments. Finally, we conclude our work in section \ref{section-conclusion}.

\section{Notations and Preliminaries}
\label{section-notation}

\subsection{Notations}

This subsection introduces some basic notations of tensor and TR decomposition. For example, a scalar, a vector, a matrix and a tensor are denoted by a normal letter $x$, a boldface lowercase letter $\mathbf{x}$, a boldface uppercase letter $\mathbf{X}$ and a calligraphic letter $\mathcal{X}$, respectively. Specifically, a $d$-order tensor of size $n_{1}\times \dotsm  \times n_{d}$ is denoted as $\mathcal{X} \in \mathbb{R}^{n_{1}\times \dotsm  \times n_{d}}$, where $n_i$ is the dimensional size corresponding to mode-$i,\;i=1,\dotsc,d$. The $\left(j_{1}\cdots j_{d}\right)$-th entry of $ \mathcal{X} $ is denoted as $x_{j_{1}\dotsm j_{d}}$. A mode-$i$ fiber of $\mathcal{X}$ is represented as $\mathbf{x}_{j_{1}\cdots j_{i-1}j_{i+1}\dotsm j_{d}}$, and a mode-$i$ slice is denoted as $\mathcal{X}_{\dotsm j_i\dotsm}$.

We use $\mathbf{E}$ to denote an identity matrix, $\mathscr{I}\left(\cdot\right)$ to denote an identity operator and $\lVert \mathbf{X} \rVert_{2}$ to denote the spectral norm of $\mathbf{X}$. The inner product of $\mathcal{X}$ and $\mathcal{Y}$ is defined as $\langle \mathcal{X},\mathcal{Y} \rangle=\sum_{j_1=1}^{n_1}\cdots \sum_{j_{d}=1}^{n_{d}}{x_{j_{1}\cdots j_{d}}y_{j_{1}\cdots j_{d}}}$. The Frobenius norm of $\mathcal{X}$ is defined as $\lVert \mathcal{X} \rVert_\text{F}=\sqrt{\langle \mathcal{X},\mathcal{X} \rangle}$. The Kronecker product and Hadamard product are expressed as $\otimes$ and $\circledast$, respectively. A zero tensor is expressed as $\mathcal{O}$. The $O\left(\cdot\right)$ is an asymptotic notation. For example, $O\left(m\right)$ means a quantity bounded in magnitude by $Cm$ for a constant $C>0$.

\subsection{Tensor Ring Decomposition}

Let $\left\{\mathcal{G}\right\}=\left\{\mathcal{G}^{\left(1\right)},\dotsc,\mathcal{G}^{\left(d\right)}\right\}$, $\mathcal{G}^{\left(i\right)} \in \mathbb{R}^{r_i\times n_i \times r_{i+1}}$ denote the normalized factors of TR decomposition on the second dimension and $\left\{\boldsymbol{\Sigma}^{\left(1\right)},\dotsc,\boldsymbol{\Sigma}^{\left(d\right)}\right\}$, $\boldsymbol{\Sigma}^{\left(i\right)}\in \mathbb{R}^{r_{i-1}\times r_i}$ denote the TR singular value matrices. The representation of TR decomposition is $x_{j_{1}\cdots j_{d}}=\operatorname{tr}\left(\boldsymbol{\Sigma}^{\left(1\right)}\mathbf{G}^{\left(1\right)}_{j_1}\cdots \boldsymbol{\Sigma}^{\left(d\right)}\mathbf{G}^{\left(d\right)}_{j_d} \right)$, where $\mathbf{G}^{\left(i\right)}_{j_i}$ is the $j_i$-th mode-$2$ slice of $\mathcal{G}^{\left(i\right)}$ and $\operatorname{tr}\left(\cdot\right) $ is the trace function. Another representation of TR decomposition is $\mathcal{X} =\sum_{t_1=1}^{r_1}\cdots \sum_{t_d=1}^{r_d}\widetilde{\mathbf{g}}^{\left(1\right)}_{t_{1}t_{2}}\circ \cdots \circ \widetilde{\mathbf{g}}^{\left(d\right)}_{t_{d}t_{1}}$, where $\widetilde{\mathbf{g}}^{\left(i\right)}_{t_{i}t_{i+1}}=\sigma^{\left(i\right)}_{t_i}\mathbf{g}^{\left(i\right)}_{t_{i}t_{i+1}}$, $\mathbf{g}^{\left(i\right)}_{t_{i}t_{i+1}}$ is the $\left(t_i,t_{i+1}\right)$-th mode-$2$ fiber of $\mathcal{G}^{\left(i\right)}$ and $\circ$ denotes the outer product.

Let $\mathbf{X}_{\left\{i,l\right\}}$ denote the $i$-shifting $l$-matricization of $\mathcal{X}$, which permutes the tensor with order $\left[i,\dotsc,d,1,\dotsc,i-1\right]$ and performs matricization along first $l$ modes. The indices of $\left(\mathbf{X}_{\left\{i,l\right\}}\right)_{pq}$ are
\begin{align*}
p=1+\sum_{k=i}^{i+l-1}{\left( j_k-1 \right)\prod_{m=i}^{k-1}{n_m}},\; q=1+\sum_{k=i+l}^{i-1}{\left( j_k-1 \right)\prod_{m=i+l}^{k-1}{n_m}}.
\end{align*}
We use $\overline{\otimes}$ to denote the TR contraction which contracts several TR factors into a new one. The formulation of TR contraction is
\begin{align*}
\left(\overline{\otimes}^{b}_{i=a}\mathcal{G}^{\left(i\right)}\right)_{t_a:t_{b+1}}=\sum^{r_a}_{t_{a+1}=1}\cdots \sum^{r_b}_{t_b=1}\mathbf{g}^{\left(a\right)}_{t_at_{a+1}} \otimes \cdots \otimes \mathbf{g}^{\left(b\right)}_{t_bt_{b+1}},
\end{align*}
where $\overline{\otimes}^{b}_{i=a}\mathcal{G}^{\left(i\right)}\in \mathbb{R}^{r_a\times \left(\prod^{b}_{i=a}n_i\right) \times r_{b+1}}$.

\section{Main Result}
\label{section-sampling}

In this section, we consider the TR completion with random sampling. The states of TR are categorized into three types: subcritical, critical and supercritical \cite{ye2018tensor}. We suppose the TR rank is $\left[r_1,\dotsc,r_d\right]$, the subcritical (supercritical) state requires $r_ir_{i+1}\leq n_i$ ($r_ir_{i+1}\geq n_i)$, $\forall i=1,\dotsc,d$, where at least one inequality is strict, and critical state requires $r_ir_{i+1}=n_i$, $\forall i=1,\dotsc,d$. We focus on a study of a (sub)critical TR since a supercritical TR can be reduced to (sub)critical by a surjective birational map \cite{ye2018tensor}. Thereafter a TR means a (sub)critical TR wherever it appears.

We use $\Omega$ to denote the set of indices of observations. Denotation $\mathscr{P}_{\Omega}$ is the orthogonal projection onto $\Omega$. We propose the following convex model for tensor completion using $i$-shifting $l$-matricization:
\begin{equation}
\label{model-balanced TR nuclear norm minimization}
\min_{\mathcal{X}} \; \sum_{i=1}^{\lceil d/2 \rceil}w_i{\lVert \mathbf{X}_{\left\{ i,l \right\}} \rVert}_*,\; \text{s. t.}\; \mathscr{P}_{\Omega}\left(\mathcal{X}\right)=\mathscr{P}_{\Omega}\left(\mathcal{T}\right).
\end{equation}

It is unlikely that a method can be guaranteed to successfully recover a tensor without the assumption of the TR factors. For instance, if tensor $\mathcal{T}\in \mathbb{R}^{n_1\times \dotsm \times n_d}$ consists of the outer product of $d$ standard basis vectors i. e., $\mathcal{T}=\mathbf{e}_{j_1}\circ \dotsm \circ \mathbf{e}_{j_d}$, in consequence $\mathcal{T}$ can not be recovered without a priori knowledge of TR factors if entry $t_{j_1\dotsm j_d}$ is not sampled. To make the recovery feasible, the TR factors are required to be not spiky. We characterize this property as the following strong TR incoherence condition, in which the mode-$2$ fibers of TR factors play a role of singular vectors (thereby we call them TR singular tensors).
\begin{lemma}[Strong TR incoherence condition]
\label{lemma-TRsip}
The tensor $\mathcal{T}\in \mathbb{R}^{n_1\times \dotsm \times n_d}$ obeys the TR strong incoherence property with parameter $\boldsymbol{\mu}=\left[\mu_1,\dotsc,\mu_d\right]$, $\boldsymbol{\mu} \succ \mathbf{0}$ if for any $i\in \left\{1,\dotsc,d\right\}$,
\begin{equation}
\label{4.2}
|\langle \mathcal{G}^{\left(i\right)}_{:j_i:},\mathcal{G}^{\left(i\right)}_{:j'_i:} \rangle -\frac{r_ir_{i+1}}{n_i}1_{j_i=j'_i}|\leq \frac{\mu_i\sqrt{r_ir_{i+1}}}{n_i},
\end{equation}
provided that the TR rank is $\left[r_1,\dotsc,r_d\right]$.
\end{lemma}
Lemma \ref{lemma-TRsip} shows that almost all tensors satisfy the strong TR incoherence property with $\mu_i=O\left(\mu_{B_i}\sqrt{\ln\left(n_i\right)}\right)$ if they obey the size property $\max\left\{ \mathcal{G}^{\left(i\right)} \right\}\leq \sqrt{\mu_{B_i}/n_i}$ with $\mu_{B_i}=O\left(1\right)$. This union bound means the TR singular tensors $\left\{ \mathbf{g}^{\left(i\right)}_{11},\dotsc,\mathbf{g}^{\left(i\right)}_{r_ir_{i+1}} \right\}$ are incoherent and there exist small values $\mu_i$ that can satisfy the strong TR incoherence property. With the only assumption about the small values of TR singular tensors, this model can generate a generic tensor with uniformly bounded TR factors, which leads to the following result.
\begin{lemma}
\label{lemma-TRunfoldingsip}
Let $\mathcal{T}\in \mathbb{R}^{n_1\times \dotsm \times n_d}$ be a fixed tensor of TR rank $\left[r_1,\dotsc,r_d\right]$ obeying the strong TR incoherence property with parameter $\boldsymbol{\mu}$, then the singular tensors of $\mathbf{X}_{\left\{i,l\right\}}$ are $\mathcal{U}^{\left\{k,l\right\}}=\prod^{k+l-1}_{i=k+1}r_i^{-1}\overline{\otimes}^{k+l-1}_{i=k}\mathcal{G}^{\left(i\right)}$ and $\mathcal{V}^{\left\{k+l,d-l\right\}}=\prod^{k-1}_{i=k+l+1}r_i^{-1}\overline{\otimes}^{k-1}_{i=k+l}\mathcal{G}^{\left(i\right)}$, and inequalities
\begin{equation*}
\left\{
\begin{aligned}
& |\langle \mathcal{U}^{\left\{k,l\right\}}_{:\bar{i}:},\mathcal{U}^{\left\{k,l\right\}}_{:\bar{j}:} \rangle-\frac{r_kr_{k+l}}{\prod^{k+l-1}_{i=k}n_i}1_{\bar{i}=\bar{j}}|\leq \frac{\mu'_{1kl}\sqrt{r_kr_{k+l}}}{\prod^{k+l-1}_{i=k}n_i} \\
& |\langle \mathcal{V}^{\left\{k+l,d-l\right\}}_{:\bar{i}:},\mathcal{V}^{\left\{k+l,d-l\right\}}_{:\bar{j}:} \rangle-\frac{r_kr_{k+l}}{\prod^{k-1}_{i=k+l}n_i}1_{\bar{i}=\bar{j}}|\leq \frac{\mu'_{2kl}\sqrt{r_kr_{k+l}}}{\prod^{k-1}_{i=k+l}n_i} \\
& |\langle \mathcal{U}^{\left\{k,l\right\}}_{:\bar{i}:},\mathcal{V}^{\left\{k+l,d-l\right\}}_{:\bar{j}:} \rangle|\leq \frac{\mu''\sqrt{r_kr_{k+l}}}{\sqrt{\prod^{d}_{i=1}n_i}}
\end{aligned}
\right.
\end{equation*}
holds with probabilities at least $1-\prod^{k+l-1}_{i=k}n_i^{-3}$, $1-\prod^{k-1}_{i=k+l}n_i^{-3}$ and $1-e^{-\frac{1}{2}\prod^{d}_{i=1}n_i}$, respectively, where 
\begin{equation}
\label{mu_kl}
\left\{
\begin{aligned}
\mu'_{1kl}=& O\left(\prod^{k+l-1}_{i=k}\mu_{B_i}\sqrt{\sum^{k+l-1}_{i=k}\ln n_i}\right),\;\mu'_{2kl}=O\left(\prod^{k-1}_{i=k+l}\mu_{B_i}\sqrt{\sum^{k-1}_{i=k+l}\ln n_i}\right) \\
\mu''=& \prod^{d}_{i=1}\mu_{B_i}
\end{aligned}
\right..
\end{equation}
\end{lemma}
Lemma \ref{lemma-TRunfoldingsip} shows that any TR unfolding $\mathbf{X}_{\left\{i,l\right\}}$ obeys the strong matrix incoherence condition if the TR is strong incoherent. Note that \cite{yu2019tensor} states that a TR unfolding obeys $\operatorname{rank}\left(\mathbf{X}_{\left\{i,l\right\}}\right)\leq r_ir_{i+l}$. We emphasize this inequality becomes equality under specific conditions. We find $\operatorname{rank}\left(\mathbf{X}_{\left\{i,l\right\}}\right)$ satisfies the equality if mode-$1$ and mode-$3$ slices are linearly independent with $l>1$ or mode-$2$ fibers are linearly independent with $l=1$. The Lemma \ref{lemma-TRunfoldingsip} will not be violated if we assume all mode-$2$ fibers are linearly independent which leads to the upper bound $\max_i \operatorname{rank}\left(\mathbf{X}_{\left\{i,l\right\}}\right)= r_ir_{i+l}$.

Now we state our main result.
\begin{theorem}[Tensor ring completion]
\label{theorem-TR sampling bound}
Under the hypothesis of Lemma \ref{lemma-TRsip}, supposing $m$ entries of $\mathcal{T}$ are observed with locations sampled uniformly at random and defining $\overline{n}_{il}:=\max\left\{\prod^{i+l-1}_{k=i}n_k,\prod^{i-1}_{k=i+l}n_k\right\}$. Then there is a numerical constant $C$ such that if
\begin{equation}
\label{bound1}
m\geq C\max_i \mu^2_{il}\overline{n}_{il}r_ir_{i+l}\ln^6\left(\overline{n}_{il}\right),
\end{equation}
$\mathcal{X}$ is the unique solution to (\ref{model-balanced TR nuclear norm minimization}) with probability at least $1-\max_i \overline{n}_{il}^{-3}$, where $\mu_{il}$ is the maximal value of (\ref{mu_kl}).
\end{theorem}
A conclusion that can be drawn directly from (\ref{mu_kl}) and (\ref{bound1}) is that on the order of $\overline{n}_{il}r_ir_{i+l}\ln^7\left(\overline{n}_{il}\right)$ samples are needed to recover $\mathcal{T}$. The bound can be improved with a suitable $l$ since a (almost) square matrix leads to lower sample complexity for completion.

As a special case, the TR unfolding can be (nearly) squared by setting $l=\lceil d/2 \rceil$ (since $\overline{n}_{i\lceil d/2 \rceil}=\inf_l \overline{n}_{il}$) if $n_i$, $i=1,\dotsc,d$ are on a same or similar order of magnitude. In this case, the tensor can be recovered with a minimal number of samples theoretically.

\section{Architecture of the Proof}
\label{section-proof}

Before we prove Theorem \ref{theorem-TR sampling bound}, we define $\mathcal{R}\triangleq \mathscr{R}\left(\left\{\mathcal{G}\right\}\right)\prod^{d}_{i=1}r_i^{-1}\sum^{\lceil d/2 \rceil}_{i=1}w_ir_ir_{i+l}$ and $\mathscr{A}^*_{il}\left(\cdot\right)$ as the tensorization operator, which is the inverse operator of $\left\{i,l\right\}$ unfolding. The following conditions are important for the proof of the main theorem (see Appendix \ref{appendix3} for details).

\begin{lemma}[Dual certificate of tensor ring completion]
\label{lemma-dual certificate}
Supposing $m$ satisfies (\ref{bound1}), then $\mathcal{X}\in \mathbb{R}^{n_1\times \dotsm \times n_d}$ is the unique minimizer to (\ref{model-balanced TR nuclear norm minimization}) if
\begin{equation}
\label{condition1}
\lVert \mathscr{P}_{T_{il}}\mathscr{P}_{\Omega_{il}}\mathscr{P}_{T_{il}}-m\prod^{d}_{i=1}n_i^{-1}\mathscr{P}_{T_{il}} \rVert_2\leq \frac{m}{2}\prod^{d}_{i=1}n_i^{-1},\; i=1,\dotsc,\lceil d/2 \rceil
\end{equation}
and there exists $\mathcal{Y}$ such that
\begin{equation}
\label{condition2}
\left\{
\begin{aligned}
& \lVert \mathscr{P}_{T_{il}}\left( \mathbf{Y}_{\left\{i,l\right\}} \right)-\mathcal{U}^{\left\{i,l\right\}}_{\left(2\right)'}\mathcal{V}^{{\left\{i+l,d-l\right\}}^{\mathrm{T}}}_{\left(2\right)} \rVert_{\mathrm{F}}\leq \frac{1}{2}\prod^{d}_{i=1}n_i^{-1} \\
& \lVert \mathscr{P}_{T_{il}^{\perp}}\left( \mathbf{Y}_{\left\{i,l\right\}} \right) \rVert_2\leq \frac{1}{2}
\end{aligned}
\right.,\; i=1,\dotsc,\lceil d/2 \rceil.
\end{equation}
\end{lemma}

\begin{proof*}
The key idea to prove Theorem \ref{theorem-TR sampling bound} is to illustrate that $\sum^{\lceil d/2 \rceil}_{i=1}w_i\lVert \mathbf{X}_{\left\{ i,l \right\}}+\Delta_{\left\{ i,l \right\}} \rVert_*>\sum^{\lceil d/2 \rceil}_{i=1}w_i\lVert \mathbf{X}_{\left\{ i,l \right\}} \rVert_*$ for any feasible perturbation $\Delta\neq \mathcal{O}$ supported in $\Omega^{\perp}$. We deduce it like follows
\begin{align*}
& \sum^{\lceil d/2 \rceil}_{i=1}w_i\lVert \mathbf{X}_{\left\{ i,l \right\}}+\boldsymbol{\Delta}_{\left\{ i,l \right\}} \rVert_*-\sum^{\lceil d/2 \rceil}_{i=1}w_i\lVert \mathbf{X}_{\left\{ i,l \right\}} \rVert_* \\
\geq& \langle \mathcal{R}+\sum^{\lceil d/2 \rceil}_{i=1}w_i\mathscr{A}^*_{il}\left(\mathbf{W}_{\left\{ i,l \right\}}\right),\Delta \rangle \\
=& \sum^{\lceil d/2 \rceil}_{i=1}w_i\lVert \mathscr{P}_{T_{il}^{\perp}}\left(\boldsymbol{\Delta}_{\left\{ i,l \right\}}\right) \rVert_*+\langle \mathcal{R}-\mathcal{Y}, \Delta \rangle \\
=& \sum^{\lceil d/2 \rceil}_{i=1}w_i\lVert \mathscr{P}_{T_{il}^{\perp}}\left(\boldsymbol{\Delta}_{\left\{ i,l \right\}}\right) \rVert_*+\sum^{\lceil d/2 \rceil}_{i=1}w_i\langle \mathbf{R}_{\left\{ i,l \right\}}-\mathscr{P}_{T_{il}}\left(\mathbf{Y}_{\left\{ i,l \right\}}\right), \mathscr{P}_{T_{il}}\left(\boldsymbol{\Delta}_{\left\{ i,l \right\}}\right) \rangle- \\
& \sum^{\lceil d/2 \rceil}_{i=1}w_i\langle \mathscr{P}_{T_{il}^{\perp}}\left(\mathbf{Y}_{\left\{ i,l \right\}}\right), \mathscr{P}_{T_{il}^{\perp}}\left(\boldsymbol{\Delta}_{\left\{ i,l \right\}}\right) \rangle \\
\geq& \sum^{\lceil d/2 \rceil}_{i=1}\frac{w_i}{2}\left(\lVert \mathscr{P}_{T_{il}^{\perp}}\left(\boldsymbol{\Delta}_{\left\{ i,l \right\}}\right) \rVert_*-\prod^{d}_{i=1}n_i^{-1}\lVert \mathscr{P}_{T_{il}}\left(\boldsymbol{\Delta}_{\left\{ i,l \right\}}\right) \rVert_{\mathrm{F}}\right) \\
\geq& 0.
\end{align*}
The first inequality comes from the convexity of nuclear norm and second-order Taylor's expansion. Since $\mathscr{P}_{T_{il}}\left(\mathbf{W}_{\left\{i,l\right\}}\right)=\mathbf{0}$, the second equality holds by choosing $\mathbf{W}_{\left\{i,l\right\}}$ such that $\langle \mathscr{A}^*_{iL}\left(\mathbf{W}_{\left\{i,L\right\}}\right),\Delta \rangle=\lVert \mathscr{P}_{T_{kL}^\perp}\left(\boldsymbol{\Delta}_{\left\{k,L\right\}}\right) \rVert_*$. The third equality is due to $\mathscr{P}_{T_{il}^{\perp}}\left(\mathbf{R}_{\left\{i,l\right\}}\right)=\mathbf{0}$ and $\mathscr{P}_{\Omega^{\perp}}\left(\mathcal{Y}\right)=\mathscr{P}_{\Omega}\left(\Delta\right)=\mathcal{O}$. The fourth inequality is because of (\ref{condition2}) and hence
\begin{equation*}
\left\{
\begin{aligned}
& \langle \mathscr{P}_{T_{il}^{\perp}}\left(\mathbf{Y}_{\left\{i,l\right\}}\right),\mathscr{P}_{T_{il}^{\perp}}\left(\boldsymbol{\Delta}_{\left\{i,l\right\}}\right) \rangle\leq \frac{1}{2}\lVert \mathscr{P}_{T_{il}^{\perp}}\left(\boldsymbol{\Delta}_{\left\{k,l\right\}}\right) \rVert_* \\
& \langle \mathbf{R}_{\left\{ i,l \right\}}-\mathscr{P}_{T_{il}}\left(\mathbf{Y}_{\left\{ i,l \right\}}\right), \mathscr{P}_{T_{il}}\left(\boldsymbol{\Delta}_{\left\{ i,l \right\}}\right) \rangle\geq -\frac{1}{2}\prod^{d}_{k=1}n_k^{-1}\lVert \mathscr{P}_{T_{il}}\left(\boldsymbol{\Delta}_{\left\{i,l\right\}}\right) \rVert_{\mathrm{F}}
\end{aligned}
\right..
\end{equation*}
The fifth inequality follows from the following deduction. Note that (\ref{condition1}) indicates
\begin{align*}
& \langle \mathscr{P}_{T_{il}}\mathscr{P}_{\Omega_{il}}\mathscr{P}_{T_{il}}\left(\boldsymbol{\Delta}_{\left\{i,l\right\}}\right)-m\prod^{d}_{k=1}n_k^{-1}\mathscr{P}_{T_{il}}\left(\boldsymbol{\Delta}_{\left\{i,l\right\}}\right), \boldsymbol{\Delta}_{\left\{k,l\right\}} \rangle \\
\geq& -\lVert \mathscr{P}_{T_{il}}\mathscr{P}_{\Omega_{il}}\mathscr{P}_{T_{il}}\left(\boldsymbol{\Delta}_{\left\{i,l\right\}}\right)-m\prod^{d}_{k=1}n_k^{-1}\mathscr{P}_{T_{il}}\left(\boldsymbol{\Delta}_{\left\{i,l\right\}}\right) \rVert_{\mathrm{F}}\lVert \Delta \rVert_{\mathrm{F}} \\
\geq& -\frac{m}{2}\sqrt{\underline{n}_{il}}\prod^{d}_{k=1}n_k^{-1}\lVert \Delta \rVert_{\mathrm{F}},
\end{align*}
where
\begin{align*}
\langle \mathscr{P}_{T_{il}}\mathscr{P}_{\Omega_{il}}\mathscr{P}_{T_{il}}\left(\boldsymbol{\Delta}_{\left\{i,l\right\}}\right), \boldsymbol{\Delta}_{\left\{i,l\right\}} \rangle=&  \lVert \mathscr{P}_{\Omega_{il}}\mathscr{P}_{T_{il}}\left(\boldsymbol{\Delta}_{\left\{i,l\right\}}\right) \rVert^2_{\mathrm{F}} \\
=& \lVert \mathscr{P}_{\Omega_{il}}\mathscr{P}_{T_{il}^{\perp}}\left(\boldsymbol{\Delta}_{\left\{k,l\right\}}\right) \rVert^2_{\mathrm{F}} \\
\leq& \lVert \mathscr{P}_{T_{il}^{\perp}}\left(\boldsymbol{\Delta}_{\left\{i,l\right\}}\right) \rVert^2_{\mathrm{F}},
\end{align*}
hence
\begin{align*}
& \lVert \mathscr{P}_{T_{il}^{\perp}}\left(\boldsymbol{\Delta}_{\left\{i,l\right\}}\right) \rVert^2_{\mathrm{F}} \\
\geq& \langle \mathscr{P}_{T_{il}}\mathscr{P}_{\Omega_{il}}\mathscr{P}_{T_{il}}\left(\boldsymbol{\Delta}_{\left\{i,l\right\}}\right), \boldsymbol{\Delta}_{\left\{i,l\right\}} \rangle \\
\geq& m\prod^{d}_{k=1}n_k^{-1}\left(\langle \mathscr{P}_{T_{il}}\left(\boldsymbol{\Delta}_{\left\{i,l\right\}}\right), \boldsymbol{\Delta}_{\left\{i,l\right\}} \rangle-\frac{1}{2}\sqrt{\underline{n}_{il}}\lVert \Delta \rVert_{\mathrm{F}}\right) \\
=& m\prod^{d}_{k=1}n_k^{-1}\left( \lVert \mathscr{P}_{T_{il}}\left(\boldsymbol{\Delta}_{\left\{i,l\right\}}\right) \rVert^2_{\mathrm{F}}-\frac{1}{2}\sqrt{\underline{n}_{il}}\sqrt{\lVert \mathscr{P}_{T_{il}}\left(\boldsymbol{\Delta}_{\left\{i,l\right\}}\right) \rVert^2_{\mathrm{F}}+\lVert \mathscr{P}_{T_{il}^{\perp}}\left(\boldsymbol{\Delta}_{\left\{i,l\right\}}\right) \rVert^2_{\mathrm{F}}}\right),
\end{align*}
where the last equality follows from the Pythagorean identity. Therefore, by writing $a\triangleq \lVert \mathscr{P}_{T_{il}}\left(\boldsymbol{\Delta}_{\left\{i,l\right\}}\right) \rVert^2_{\mathrm{F}}$ and $b\triangleq \lVert \mathscr{P}_{T_{il}^{\perp}}\left(\boldsymbol{\Delta}_{\left\{i,l\right\}}\right) \rVert^2_{\mathrm{F}}$ we have the quadratic inequality
\begin{align*}
a^2-\left(\frac{2}{m}\prod^{d}_{k=1}n_k b+\frac{\underline{n}_{kl}}{4}\right)a+\frac{1}{m^2}\prod^{d}_{k=1}n_k^2 b^2-\frac{\underline{n}_{il}}{4}b\leq 0
\end{align*}
whose discriminant follows $\triangle=\left(1+m^{-1}\prod^{d}_{k=1}n_k \right)b+\underline{n}_{il}^2/16>0$. Then
\begin{align*}
a\leq& \frac{1}{m}\prod^{d}_{k=1}n_k b+\frac{\underline{n}_{il}}{8}+\frac{1}{2}\sqrt{\triangle} \\
\leq& \frac{1}{m}\prod^{d}_{k=1}n_k b+\frac{\underline{n}_{il}}{4}+\left(1+\frac{1}{m}\prod^{d}_{k=1}n_k\right)\frac{b}{\underline{n}_{il}},
\end{align*}
which leads to
\begin{align*}
\lVert \mathscr{P}_{T_{il}^{\perp}}\left(\boldsymbol{\Delta}_{\left\{i,l\right\}}\right) \rVert_*\geq \lVert \mathscr{P}_{T_{il}^{\perp}}\left(\boldsymbol{\Delta}_{\left\{i,l\right\}}\right) \rVert_{\mathrm{F}}=\sqrt{b}\geq& \sqrt{\frac{m\prod^{d}_{k=1}n_k^{-1}\underline{n}_{il}\left(a-\frac{1}{4}\underline{n}_{il}\right)}{m\prod^{d}_{k=1}n_k^{-1}+\underline{n}_{il}+1}} \\
\geq& \sqrt{\prod^{d}_{k=1}n_k^{-2}a} \\
=& \prod^{d}_{k=1}n_k^{-1}\lVert \mathscr{P}_{T_{il}}\left(\boldsymbol{\Delta}_{\left\{i,l\right\}}\right) \rVert_{\mathrm{F}}.
\end{align*}

Hence, we prove for any $\Delta\neq \mathcal{O}$, $\mathscr{P}_{\Omega}\left(\Delta\right)=\mathcal{O}$ there is $\sum^{\lceil d/2 \rceil}_{i=1}w_i\lVert \mathbf{X}_{\left\{ i,l \right\}}+\Delta_{\left\{ i,l \right\}} \rVert_*>\sum^{\lceil d/2 \rceil}_{i=1}w_i\lVert \mathbf{X}_{\left\{ i,l \right\}} \rVert_*$, which indicates the uniqueness of the minimizer of (\ref{model-balanced TR nuclear norm minimization}). End of proof. $\hfill \blacksquare$
\end{proof*}

\section{Numerical Experiments}
\label{section-experiment}

In this section, three groups of datasets are used for tensor completion experiments, i.e., synthetic data, real-world images and videos. To illustrate the practical applicability of our model for tensor completion, the proposed model (\ref{model-balanced TR nuclear norm minimization}) is solved by alternating direction method of multipliers (ADMM) \cite{boyd2011distributed}. The corresponding algorithm is called tensor completion via tensor ring with balanced unfolding (TRBU). In each iteration of TRBU, the penalty parameter $\mu$ satisfies $\mu^k=\beta \mu^{k-1}$ with $\beta\in \left(0,2\right)$ \cite{nishihara2015general}.

Eight algorithms are benchmarked on real-world data, including tensor ring nuclear norm minimization for tensor completion (TRNNM) \cite{yu2019tensor}, low rank tensor completion via alternating least square (TR-ALS) \cite{wang2017efficient}, simple low rank tensor completion via tensor train (SiLRTC-TT) \cite{bengua2017efficient}, high accuracy low rank tensor completion algorithm (HaLRTC) \cite{liu2013tensor}, low rank tensor completion via tensor nuclear norm minimization (LRTC-TNN) \cite{lu2016libadmm}, Bayesian CP Factorization (FBCP) for image recovery \cite{zhao2015bayesian}, smooth low rank tensor tree completion (STTC) \cite{liu2019image} and the proposed one. These methods are based on different tensor decompositions, including  CP, Tucker, t-SVD, HT, TT and TR decompositions. Table \ref{comparison-complexity} shows the algorithmic complexity of eight algorithms, where $d$ is the tensor dimension, $n$ is the dimensional size, $m$ is the number of samples and $r$ ($\left[r,\dotsc,r\right]$) is the tensor rank corresponding to each tensor decomposition.
\begin{table*}[htbp]
\centering
\caption{Comparison of complexity of eight algorithms in one iteration.}
\label{comparison-complexity}
\begin{tabular}{ccccc}
\toprule
Algorithm & TRBU & TRNNM & TR-ALS & SiLRTC-TT \\
\midrule 
Complexity & $O\left(dn^{3d/2}\right)$ & $O\left(dn^{3d/2}\right)$ & $O\left(dmr^4\right)$ & $O\left(dn^{3d/2}\right)$ \\
\midrule 
Algorithm & LRTC-TNN & FBCP & HaLRTC & STTC \\
\midrule 
Complexity & $O\left(n^{d+1}\right)$ & $O\left(dmr^2\right)$ & $O\left(dn^{3d-3}\right)$ & $O\left(dn^{d+1}\right)$ \\
\bottomrule
\end{tabular}
\end{table*}

There are several metrics for evaluating the recovery quality of visual data. The relative error (RE) is defined as $\text{RE}=\lVert \hat{\mathcal{X}}-\mathcal{X} \rVert_{\mathrm{F}}/{\lVert \mathcal{X} \rVert}_{\mathrm{F}}$, where $\mathcal{X}$ is the ground truth and $\hat{\mathcal{X}}$ is the estimate of $\mathcal{X}$. The peak signal-to-noise ratio (PSNR) is a ratio between the maximum possible power of a signal and the power of corrupting noise \cite{barnsley1993fractal}. We use computational CPU time (in seconds) as a measure of algorithmic complexity.

The sampling rate (SR) is defined as the ratio of the number of samples to the total number of the elements of tensor $\mathcal{X}$, which is denoted as $\text{SR}=|\mathbb{O}|/|\mathcal{X}|$. For fair comparison, the parameters in each algorithm are tuned to give optimal performance. For the proposed TRBU algorithm, one of the stop criteria is that the relative change $\text{RC}=\lVert \mathcal{X}_k-\mathcal{X}_{k-1}\rVert_{\mathrm{F}}/\lVert \mathcal{X}_{k-1}\rVert_{\mathrm{F}}$ is less than a tolerance we set to $1\times 10^{-8}$. We set the maximal number of iterations $K=500$ in experiments on synthetic data and $K=100$ in experiments on real-world data.

In the remainder of this section, we verify the theoretic analysis using synthetic data. The real-world data is also employed to test the proposed method, including images and videos. All the experiments are conducted in MATLAB 9.3.0 on a computer with a 2.8GHz CPU of Intel Core i7 and a 16GB RAM.

\subsection{Exact Recovery from Random Problem}

To testify Theorem \ref{theorem-TR sampling bound}, we generated two tensors in the first group of experiments: (a) a $8$-order tensor $\mathcal{X} \in \mathbb{R}^{3\times\dotsm\times3}$ of TR rank $\left[2,\dotsc,2\right]$; (b) a $6$-order tensor $\mathcal{X} \in \mathbb{R}^{6\times\dotsm\times6}$ of TR rank $\left[3,\dotsc,3\right]$. The entries of TR factors are independently sampled from the normal distribution $\mathcal{N}\left(0,1/\sqrt{n}\right)$. Their sampling rates range from $5\%$ to $95\%$ with linear interval $5\%$. For each tensor with different sampling rates, we run the TRBU algorithm $100$ times to recover its $d/2$ unfolding matrices, i.e., $\mathbf{X}_{\left\{1,l\right\}},\;l=1,\dotsc,d/2$. The parameter setting for proposed TRBU are $\beta=1.028$ and $\mu_0=10^{-2.5}$.

\begin{figure}[htbp]
\centering
\begin{subfigure}[t]{0.45\textwidth}
\centering
\includegraphics[scale=0.15]{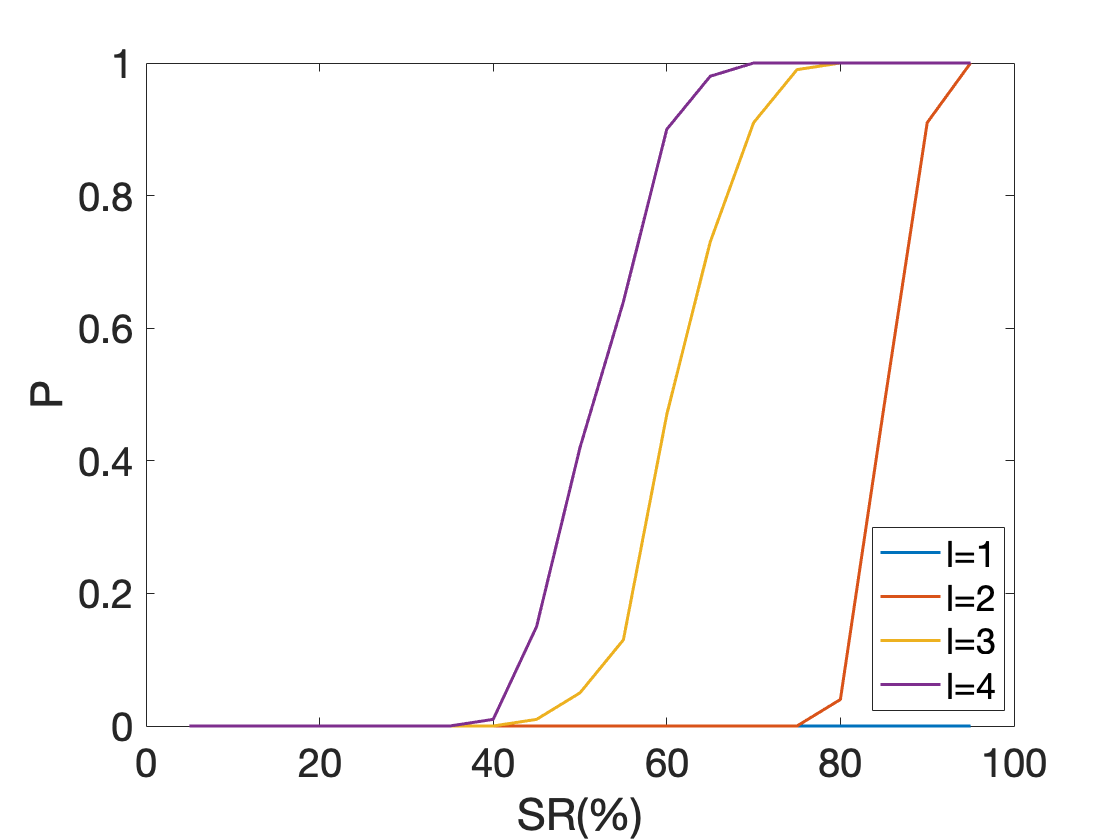}
\subcaption{An $8$-order tensor $\mathcal{X}\in \mathbb{R}^{3\times \dotsm \times3}$ of TR rank $\left(2,\dotsc,2\right)$.}
\end{subfigure}
\qquad
\begin{subfigure}[t]{0.45\textwidth}
\centering
\includegraphics[scale=0.15]{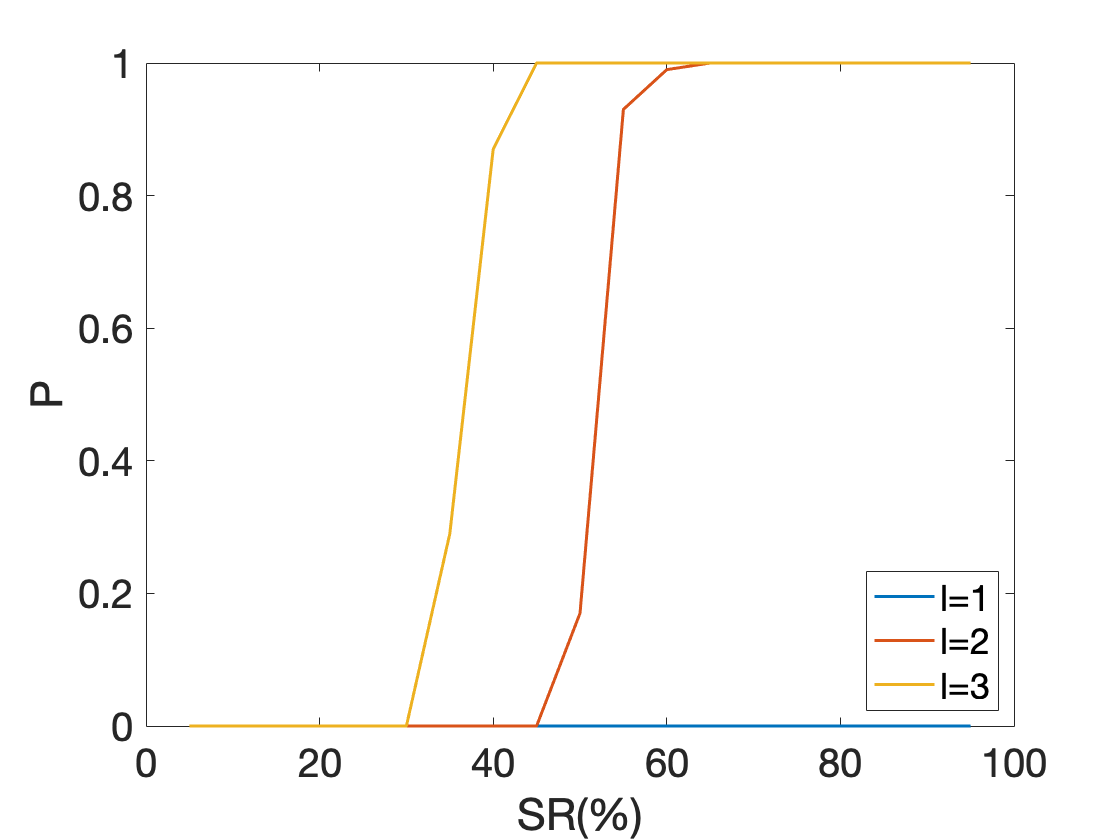}
\subcaption{A $6$-order tensor $\mathcal{X}\in \mathbb{R}^{6\times \dotsm \times6}$ of TR rank $\left(3,\dotsc,3\right)$.}
\end{subfigure}
\caption{Three experiments on randomly generated tensors.}
\label{result-synthetic data1}
\end{figure}

The averaged results are shown in Fig. \ref{result-synthetic data1}, which gives the recovery probabilities with respect to various sampling rates. In this experiment, a recovery is considered to be successful if $\text{RE}<1\times 10^{-6}$. It can be seen from Fig. \ref{result-synthetic data1} that a balanced unfolding matrix is easier to recover than an unbalanced one, which validates our claim that the more balance the matrix is, the easier it is to recover. Due to the superiority of TRBU when step length $l=\lceil d/2 \rceil$, we fix $l=\lceil d/2 \rceil$ in default in our later experiments.

In the second group of experiments, we generated two tensors, one with $d=5$ and $n=12$, the other with $d=4$ and $n=20$. The TR ranks are $\left[0.2n,\dotsc,0.2n\right]$ and $\left[0.3n,\dotsc,0.3n\right]$, respectively. The sampling rate is $0.1$, $0.2$, $0.3$ and $0.4$. We run the algorithm $10$ times for each parameter setting. We set $\beta=1.028$ and $\mu_0=10^{-2.5}$ for TRBU algorithm in this experiment.

Table \ref{result-synthetic data2} reports the recovery results of four scenarios. We examine the TR rank of the recovered tensor by computing $\operatorname{rank}\left(\hat{\mathbf{L}}_{\left\{i,l\right\}}\right)$, $i=1,\dotsc,d$, $l=1,\dotsc,d-1$ and checking if they are equal to the square of the value of the pre-defined TR rank. In all cases, the relative error $\lVert \hat{\mathcal{L}}-\mathcal{L}_0 \rVert_{\mathrm{F}}/\lVert \mathcal{L}_0 \rVert_{\mathrm{F}}$ is less than $1\times 10^{-4}$. Moreover, the TR ranks of the recovered tensors are consistent with the pre-defined ones, which illustrates the effectiveness of the proposed algorithm.

\begin{table*}
\centering
\caption{Correct recovery results of a variety of randomly generated problems.}
\label{result-synthetic data2}
\begin{tabular}{c|c|c|c|c|c|c}
\toprule
Size $n$ & Order $d$ & \#samples $m$ & $\operatorname{rank}\left(\mathcal{L}_0\right)$ & $\operatorname{rank}\left(\hat{\mathcal{L}}\right)$ & $\frac{\lVert \hat{\mathcal{L}}-\mathcal{L}_0 \rVert_{\mathrm{F}}}{\lVert \mathcal{L}_0 \rVert_{\mathrm{F}}}$ & CPU time (s) \\
\hline \hline
\multirow{2}{*}{12} & \multirow{2}{*}{5} & 24884 & 2 & 2 & $5.27\times 10^{-6}$ & $1.26\times 10^{1}$ \\
\cline{3-7}
& & 49767 & 2 & 2 & $5.38\times 10^{-8}$ & $1.16\times 10^{1}$ \\
\hline
\multirow{2}{*}{20} & \multirow{2}{*}{4} & 16000 & 4 & 4 & $2.32\times 10^{-5}$ & $1.41\times 10^{1}$ \\
\cline{3-7}
& & 32000 & 4 & 4 & $6.31\times 10^{-8}$ & $1.42\times 10^{1}$ \\
\hline
\multicolumn{7}{c}{$\operatorname{rank}\left(\mathcal{L}_0\right)=\lfloor 0.2n \rfloor\;(m=\lceil 0.1n^d \rceil\;and\;\lceil 0.2n^d \rceil)$} \\
\multicolumn{7}{c}{} \\

\hline
\multirow{2}{*}{12} & \multirow{2}{*}{5} & 24884 & 3 & 3 & $3.17\times 10^{-6}$ & $1.45\times 10^{1}$ \\
\cline{3-7}
& & 49767 & 3 & 3 & $2.27\times 10^{-7}$ & $1.54\times 10^{1}$ \\
\hline
\multirow{2}{*}{20} & \multirow{2}{*}{4} & 16000 & 6 & 6 & $6.83\times 10^{-5}$ & $2.11\times 10^{1}$ \\
\cline{3-7}
& & 32000 & 6 & 6 & $6.88\times 10^{-7}$ & $2.04\times 10^{1}$ \\
\hline
\multicolumn{7}{c}{$\operatorname{rank}\left(\mathcal{L}_0\right)=\lfloor 0.3n \rfloor\;(m=\lceil 0.1n^d \rceil\;and\;\lceil 0.2n^d \rceil)$} \\
\multicolumn{7}{c}{} \\

\hline
\multirow{2}{*}{12} & \multirow{2}{*}{5} & 74650 & 2 & 2 & $7.70\times 10^{-8}$ & $9.02\times 10^{0}$ \\
\cline{3-7}
& & 99533 & 2 & 2 & $2.20\times 10^{-8}$ & $6.30\times 10^{0}$ \\
\hline
\multirow{2}{*}{20} & \multirow{2}{*}{4} & 48000 & 4 & 4 & $2.92\times 10^{-8}$ & $1.01\times 10^{1}$ \\
\cline{3-7}
& & 64000 & 4 & 4 & $2.21\times 10^{-8}$ & $7.47\times 10^{0}$ \\
\hline
\multicolumn{7}{c}{$\operatorname{rank}\left(\mathcal{L}_0\right)=\lfloor 0.2n \rfloor\;(m=\lceil 0.3n^d \rceil\;and\;\lceil 0.4n^d \rceil)$} \\
\multicolumn{7}{c}{} \\

\hline
\multirow{2}{*}{12} & \multirow{2}{*}{5} & 74650 & 3 & 3 & $5.17\times 10^{-8}$ & $8.62\times 10^{0}$ \\
\cline{3-7}
& & 99533 & 3 & 3 & $1.56\times 10^{-8}$ & $5.64\times 10^{0}$ \\
\hline
\multirow{2}{*}{20} & \multirow{2}{*}{4} & 48000 & 6 & 6 & $4.41\times 10^{-7}$ & $1.95\times 10^{1}$ \\
\cline{3-7}
& & 64000 & 6 & 6 & $1.64\times 10^{-8}$ & $1.27\times 10^{1}$ \\
\hline
\multicolumn{7}{c}{$\operatorname{rank}\left(\mathcal{L}_0\right)=\lfloor 0.3n \rfloor\;(m=\lceil 0.3n^d \rceil\;and\;\lceil 0.4n^d \rceil)$} \\
\bottomrule
\end{tabular}
\end{table*}

\subsection{Phase Transition in TR rank with Varying Sampling Rates}

\begin{figure}[htbp]
\centering
\begin{subfigure}[t]{0.3\textwidth}
\centering
\includegraphics[scale=0.12]{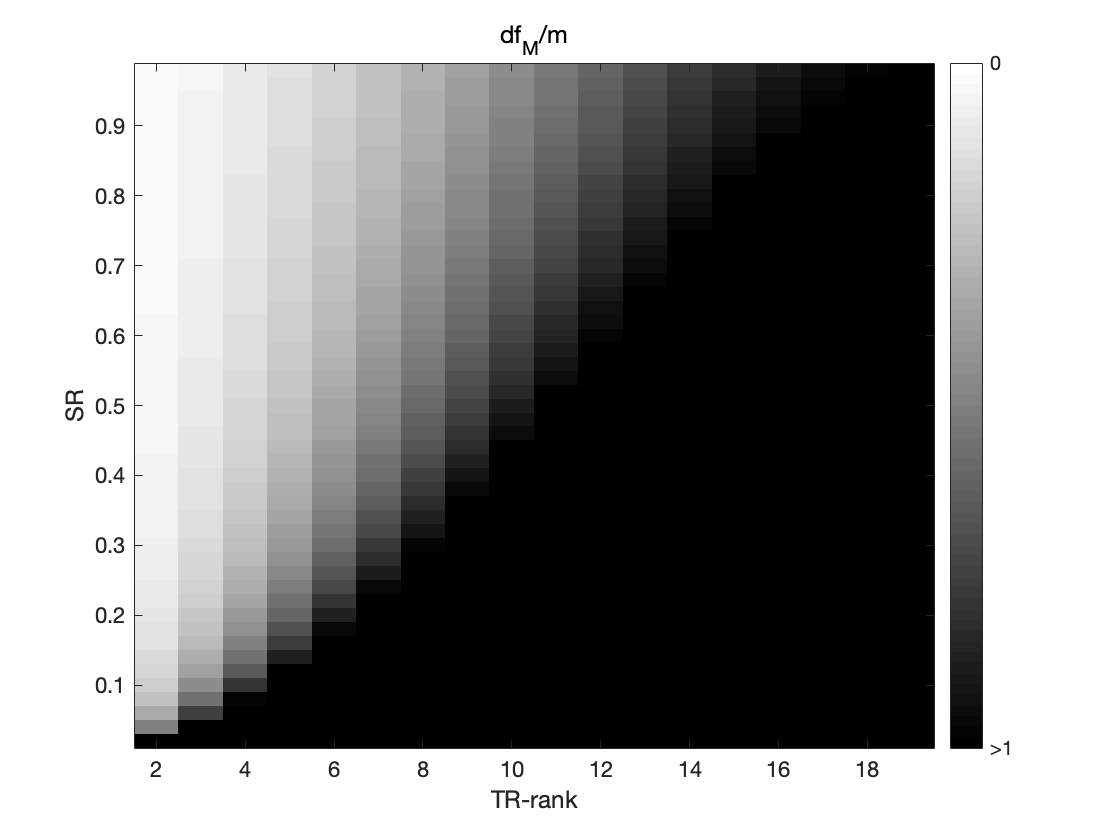}
\subcaption{The ratio of degree of freedom of a square TR unfolding to the number of samples.}
\end{subfigure}
\;
\begin{subfigure}[t]{0.3\textwidth}
\centering
\includegraphics[scale=0.12]{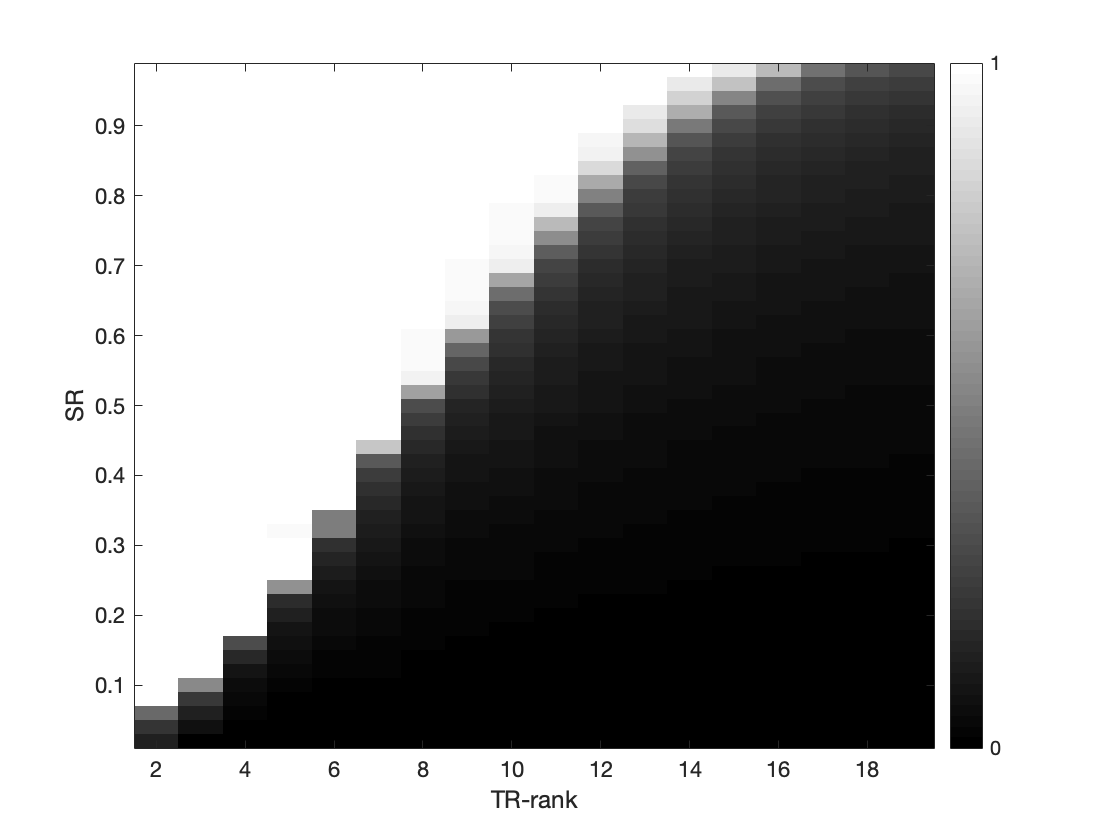}
\subcaption{The phase transition in TR rank with varying sampling rates.}
\end{subfigure}
\;
\begin{subfigure}[t]{0.3\textwidth}
\centering
\includegraphics[scale=0.12]{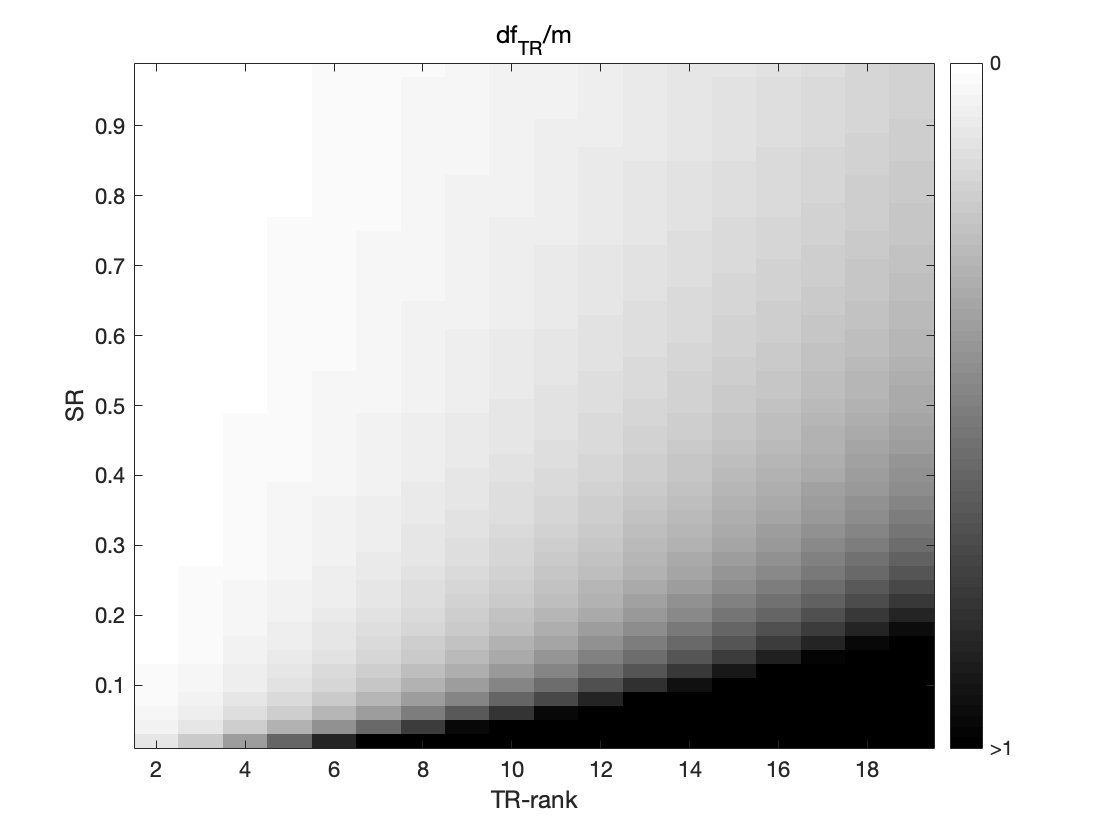}
\subcaption{The ratio of degree of freedom of a TR to the number of samples.}
\end{subfigure}
\caption{The completion result of a $4$-order tensor $\mathcal{X}\in \mathbb{R}^{20\times \dotsm \times 20}$ under various settings of TR ranks and sampling rates.}
\label{result-synthetic data3}
\end{figure}

In order to verify the recovery guarantee in Theorem \ref{theorem-TR sampling bound}, we generated a $4$-order tensor $\mathcal{X}\in \mathbb{R}^{20\times \dotsm \times 20}$ by contracting independent TR factors whose entries are sampled from i.i.d. $\mathcal{N}\left(0,1/\sqrt{20}\right)$ distributions. Theoretically, this tensor can be recovered successfully when $\text{df}_{\text{M}}/m<C$, where $\text{df}_{\text{M}}=r^2\left(2\sqrt{n^d}-r^2\right)$ is the degree of freedom (df) of a square unfolding and $C=O\left(1\right)$ is a constant. The $\text{df}_{\text{M}}$ changing with sampling rate and TR rank are drawn in Fig. \ref{result-synthetic data3}(a). The color of each cell represents the value of $\text{df}_{\text{M}}$.

We vary TR rank from $2$ to $19$ to ensure $\text{df}_{\text{M}}$ is positive. Then we carried out $10$ experiments for each $\left(\text{SR},\text{TR rank}\right)$ pair, where the algorithmic parameters $\beta$ and $\mu_0$ are set to $1.028$ and $10^{-2}$, respectively. For each experiment, the recovery is considered to be successful if the relative error is less than $1\times 10^{-4}$. The phase transition of the tensor completion is shown in Fig. \ref{result-synthetic data3}(b), where the color bar reflects the empirical recovery rate which is scaled between $0$ and $1$. A white patch indicates a success of all experiments, while a black one represents a failure in all experiments.

The results show similar boundaries in Fig. \ref{result-synthetic data3}(a) and Fig. \ref{result-synthetic data3}(b), which is a validation for our main theory.  As a comparison, the degree of freedom of the TR $\text{df}_{\text{TR}}=dnr^2-dr^2+1$ \cite{ye2018tensor} is plotted in Fig. \ref{result-synthetic data3}(c), which suggests the sampling bound in Theorem \ref{theorem-TR sampling bound} may be improved in some way since the tensor cannot be recovered in the area where $\text{df}_{\text{M}}/m>C$ and $\text{df}_{\text{TR}}/m<C$.

\subsection{Color Images}

The visual data tensorization (VDT) \cite{latorre2005image, bengua2017efficient} transforms an image into a real ket of a Hilbert space by an appropriate block structured addressing. For an image of size $M\times N\times 3$,  VDT first reshapes it into a tensor of size $m_1\times\dotsm\times m_K\times n_1\dotsm\times n_K\times 3$, then permutes and reshapes the resulting tensor into another one with size $m_1n_1\times\dotsm\times m_Kn_K\times 3$.

\begin{figure}[htbp]
\centering
\begin{subfigure}[t]{0.1\textwidth}
\centering
\includegraphics[scale=0.15]{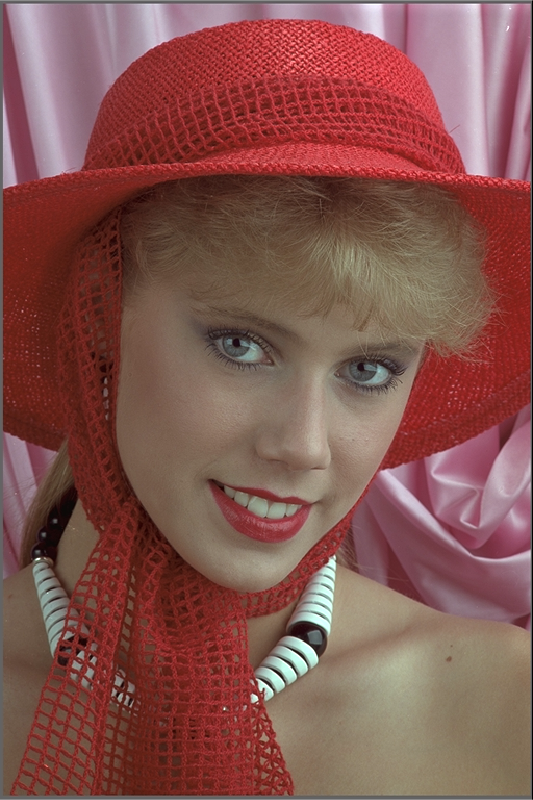}
\subcaption*{\emph{kodim04}}
\end{subfigure}
\begin{subfigure}[t]{0.1\textwidth}
\centering
\includegraphics[scale=0.3]{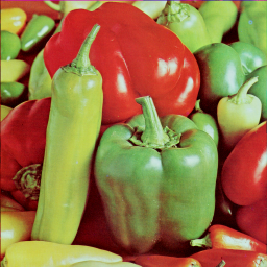}
\subcaption*{\emph{peppers}}
\end{subfigure}
\begin{subfigure}[t]{0.1\textwidth}
\centering
\includegraphics[scale=0.15]{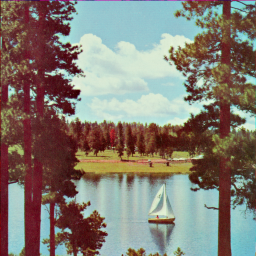}
\subcaption*{\emph{sailboat}}
\end{subfigure}
\begin{subfigure}[t]{0.1\textwidth}
\centering
\includegraphics[scale=0.3]{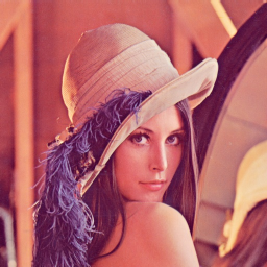}
\subcaption*{\emph{lena}}
\end{subfigure}
\begin{subfigure}[t]{0.1\textwidth}
\centering
\includegraphics[scale=0.15]{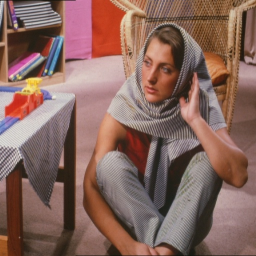}
\subcaption*{\emph{barbara}}
\end{subfigure}
\begin{subfigure}[t]{0.1\textwidth}
\centering
\includegraphics[scale=0.3]{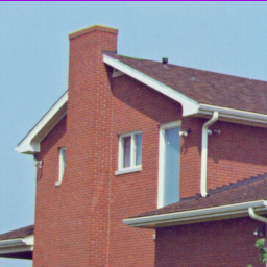}
\subcaption*{\emph{house}}
\end{subfigure}
\begin{subfigure}[t]{0.1\textwidth}
\centering
\includegraphics[scale=0.3]{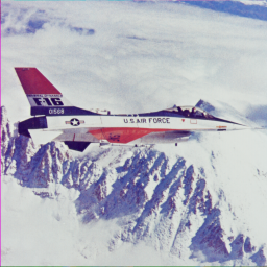}
\subcaption*{\emph{airplane}}
\end{subfigure}
\begin{subfigure}[t]{0.1\textwidth}
\centering
\includegraphics[scale=0.13]{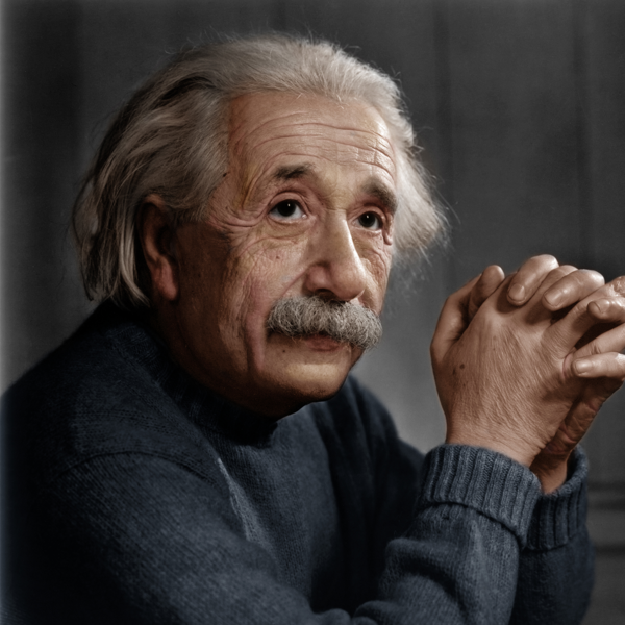}
\subcaption*{\emph{Einstein}}
\end{subfigure}
\caption{The original copies of eight images.}
\label{original-image}
\end{figure}

Eight RGB images are used in the first group of experiments, including \emph{kodim04} \footnote{http://r0k.us/graphics/kodak/kodim04.html}, \emph{peppers}, \emph{sailboat}, \emph{lena}, \emph{barbara}, \emph{house}, \emph{airplane} and \emph{Einstein} \cite{wang2017efficient}. The original images are shown in Fig. \ref{original-image}.
To perform the experiments, we first apply VDT to these images. Specifically, we reshape \emph{kodim04} by $\left[2*\operatorname{ones}\left(1,8\right),3,2*\operatorname{ones}\left(1,9\right),3\right]$ and get a tensor of size $\left[4*\operatorname{ones}\left(1,8\right),6,3\right]$. We reshape \emph{Einstein} by $\left[2,2,2,3,5,5,5,5,3,2,2,2,3\right]$ and derive a tensor of size $\left[10,10,6,6,10,10,3\right]$. For other images, we reshape them into $17$-order tensors of size $2\times\dotsm\times2\times3$, further they are reshaped into $9$-order tensors of size $4\times\dotsm\times4\times3$. Note that the VDT is by manual operation and the result can change with the choice.

After the VDT operation, we compare the proposed algorithm with the state-of-the-art ones. The FBCP method needs a pre-defined maximal CP rank that is very time-consuming. Specifically, the maximum CP ranks are $50$ for \emph{kodim04}, $60$ for \emph{Einstein} and $100$ for other images, otherwise the machine will be out of memory. The TR rank of all images is $14$ for TR-ALS due to the computational source limit. The sampling rates for all images are from $10\%$ to $90\%$. For each image with different sampling rate, we conducted $10$ experiments, where the parameter setting are $\beta=1.028$ and $\mu_0=1\times10^{-3.7}$.

\begin{figure}[htbp]
\centering
\begin{subfigure}[t]{0.45\textwidth}
\centering
\includegraphics[scale=0.17]{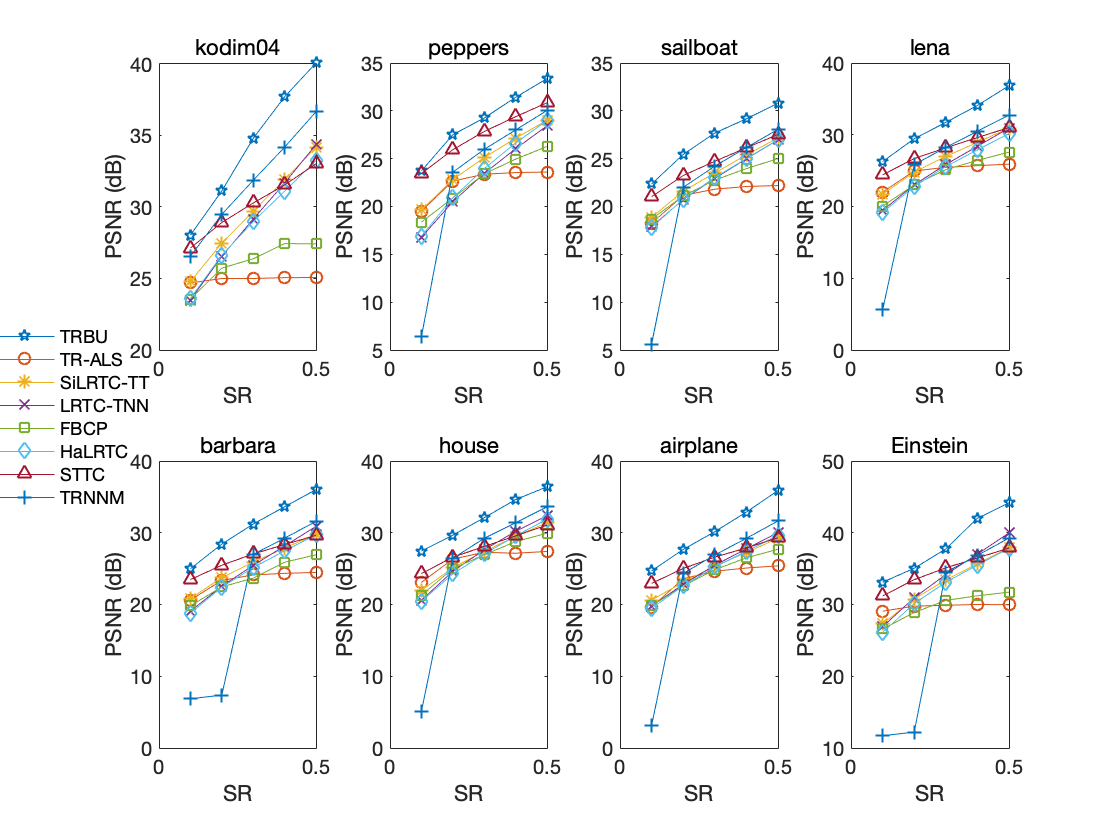}
\subcaption{Comparison of eight algorithms (with VDT operation) on PSNR (in dB).}
\end{subfigure}
\begin{subfigure}[t]{0.45\textwidth}
\centering
\includegraphics[scale=0.17]{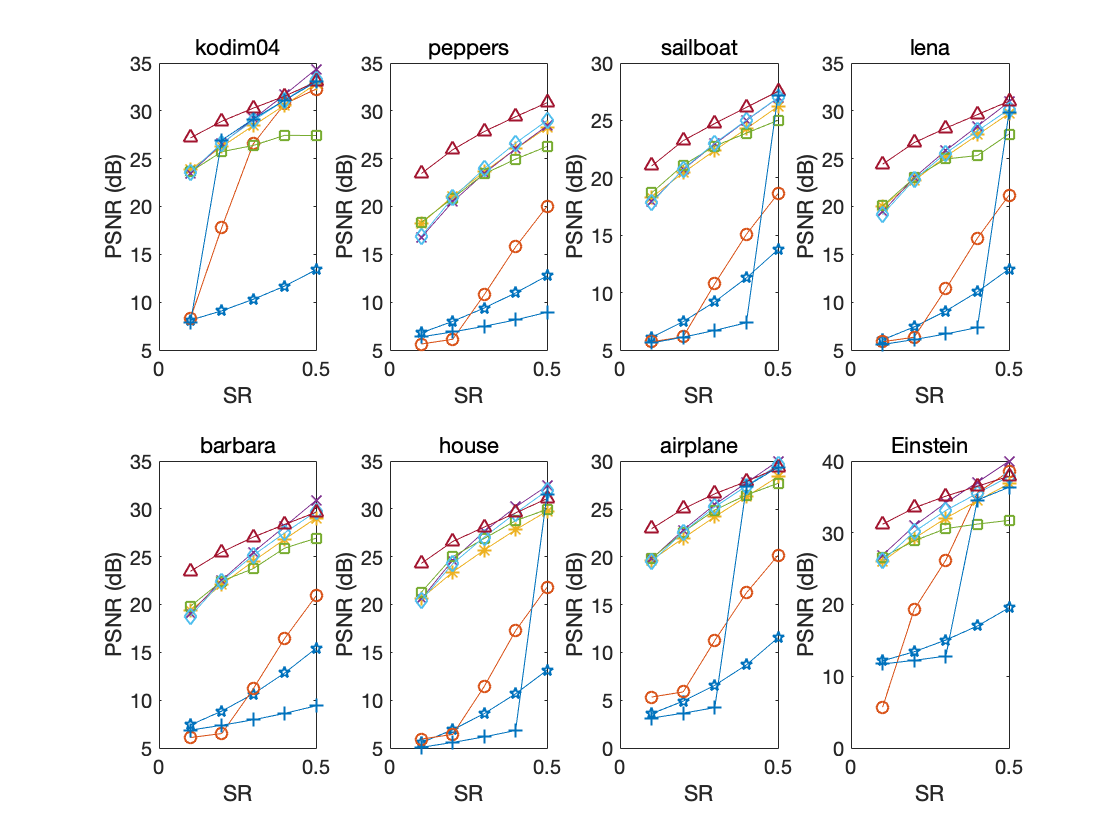}
\subcaption{Comparison of eight algorithms (without VDT operation) on PSNR (in dB).}
\end{subfigure}

\begin{subfigure}[t]{0.45\textwidth}
\centering
\includegraphics[scale=0.17]{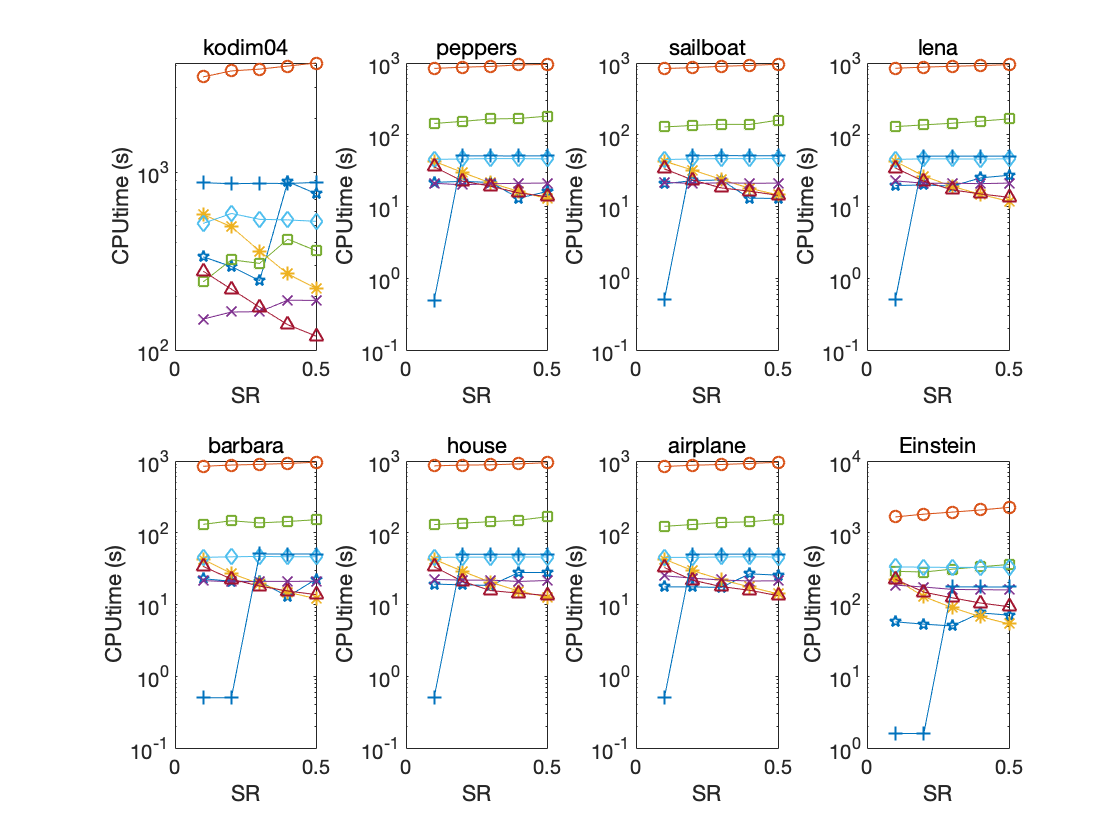}
\subcaption{Comparison of eight algorithms (with VDT operation) on CPU time (in seconds).}
\end{subfigure}
\begin{subfigure}[t]{0.45\textwidth}
\centering
\includegraphics[scale=0.17]{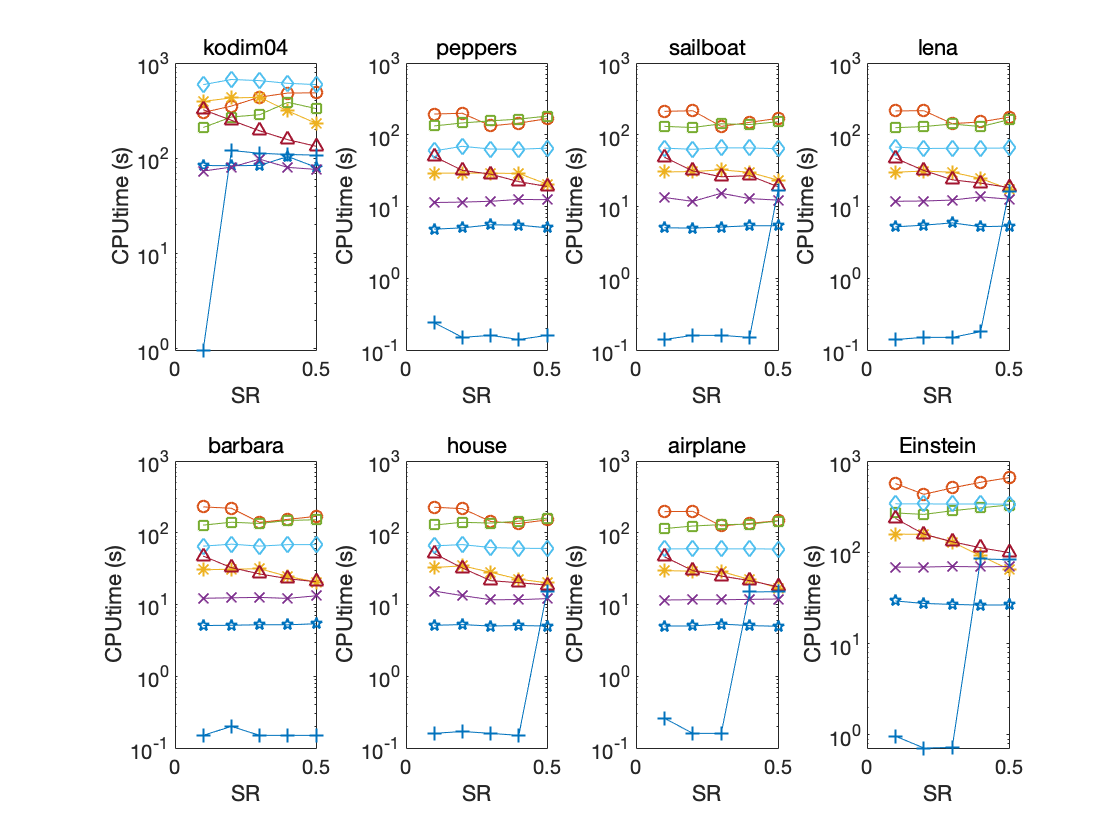}
\subcaption{Comparison of eight algorithms (without VDT operation) on CPU time (in seconds).}
\end{subfigure}
\caption{Recovery result of eight images based on eight algorithms.}
\label{result-image1}
\end{figure}

As shown in Fig. \ref{result-image1}, we compared the performance of eight algorithms both on completing the original low-order tensors and the high-order tensors from VDT processing. First, the performance of FBCP, HaLRTC, STTC and LRTC-TNN is very close in low order and high order cases by comparing Fig. \ref{result-image1}(a) and Fig. \ref{result-image1}(b). In addition, the TRBU, TRNNM, TR-ALS and SiLRTC-TT with VDT operation perform better than these without VDT, which shows that the TR decomposition and TT decomposition based methods are more suitable for solving high-order data completion problems. Moreover, TR-ALS is the most time-consuming of all algorithms in all experiments. Since the TR rank is fixed, its performance does not improve as the sampling rate increases in high order case and we suspect the model is over-fitted \cite{wang2017efficient}, while an under-fitting problem occurs in low order case. When completing high-dimensional data, TRBU is superior to other algorithms in terms of PSNR, which shows the effectiveness of TRBU in the case of recovering high-order tensors.

To simulate the non-uniform sampling situation, we use two RGB images in the second group of experiments, namely \emph{house} and \emph{llama}. The maximal CP rank for FBCP is $100$. The TR rank used in TR-ALS is $14$. We set $\beta=1.028$ and $\mu_0=10^{-3}$ in this experiment.

\begin{figure}[htbp]
\centering
\begin{subfigure}[t]{0.1\textwidth}
\centering
\subcaption*{Original}
\includegraphics[scale=0.3]{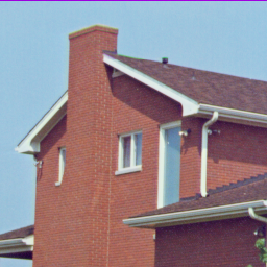}
\end{subfigure}
\begin{subfigure}[t]{0.1\textwidth}
\centering
\subcaption*{Observed}
\includegraphics[scale=0.3]{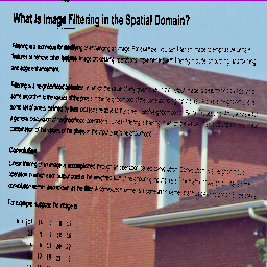}
\end{subfigure}

\begin{subfigure}[t]{0.1\textwidth}
\centering
\subcaption*{$41.936$ dB}
\includegraphics[scale=0.3]{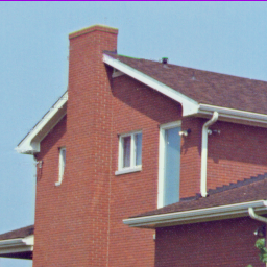}
\end{subfigure}
\begin{subfigure}[t]{0.1\textwidth}
\centering
\subcaption*{$27.506$ dB}
\includegraphics[scale=0.3]{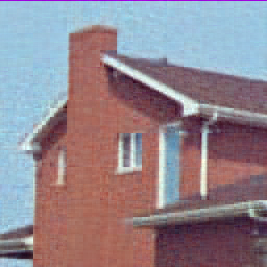}
\end{subfigure}
\begin{subfigure}[t]{0.1\textwidth}
\centering
\subcaption*{$39.146$ dB}
\includegraphics[scale=0.3]{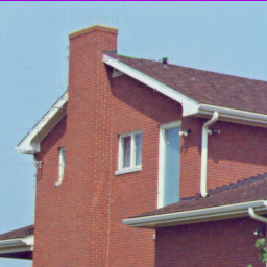}
\end{subfigure}
\begin{subfigure}[t]{0.1\textwidth}
\centering
\subcaption*{$38.485$ dB}
\includegraphics[scale=0.3]{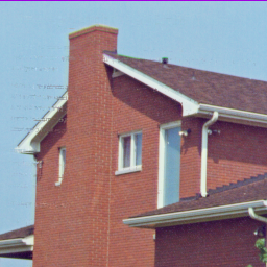}
\end{subfigure}
\begin{subfigure}[t]{0.1\textwidth}
\centering
\subcaption*{$34.373$ dB}
\includegraphics[scale=0.3]{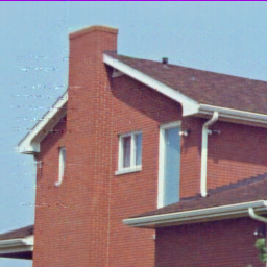}
\end{subfigure}
\begin{subfigure}[t]{0.1\textwidth}
\centering
\subcaption*{$38.562$ dB}
\includegraphics[scale=0.3]{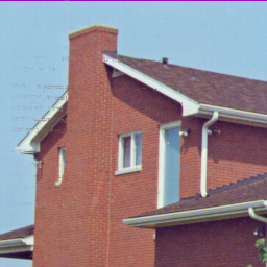}
\end{subfigure}
\begin{subfigure}[t]{0.1\textwidth}
\centering
\subcaption*{$40.071$ dB}
\includegraphics[scale=0.3]{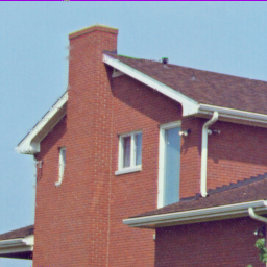}
\end{subfigure}
\begin{subfigure}[t]{0.1\textwidth}
\centering
\subcaption*{$40.336$ dB}
\includegraphics[scale=0.3]{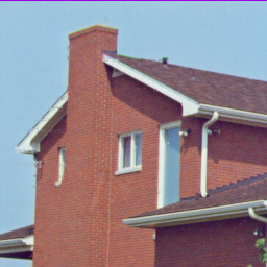}
\end{subfigure}

\begin{subfigure}[t]{0.1\textwidth}
\centering
\subcaption*{$15.459$ dB}
\includegraphics[scale=0.3]{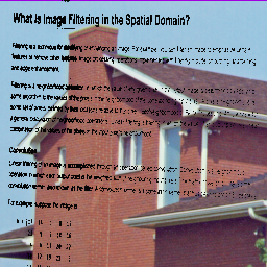}
\subcaption*{TRBU}
\end{subfigure}
\begin{subfigure}[t]{0.1\textwidth}
\centering
\subcaption*{$32.676$ dB}
\includegraphics[scale=0.3]{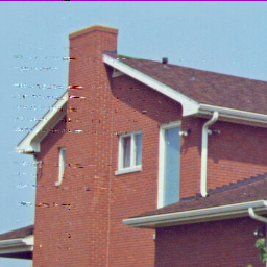}
\subcaption*{TR-ALS}
\end{subfigure}
\begin{subfigure}[t]{0.1\textwidth}
\centering
\subcaption*{$37.961$ dB}
\includegraphics[scale=0.3]{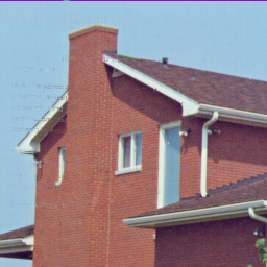}
\subcaption*{SiLRTC-TT}
\end{subfigure}
\begin{subfigure}[t]{0.1\textwidth}
\centering
\subcaption*{$38.486$ dB}
\includegraphics[scale=0.3]{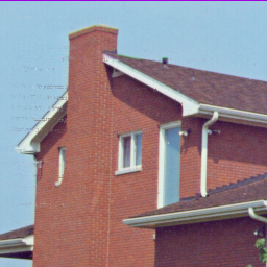}
\subcaption*{LRTC-TNN}
\end{subfigure}
\begin{subfigure}[t]{0.1\textwidth}
\centering
\subcaption*{$34.373$ dB}
\includegraphics[scale=0.3]{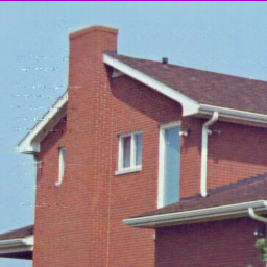}
\subcaption*{FBCP}
\end{subfigure}
\begin{subfigure}[t]{0.1\textwidth}
\centering
\subcaption*{$38.562$ dB}
\includegraphics[scale=0.3]{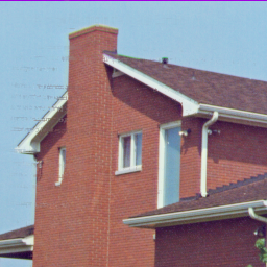}
\subcaption*{HaLRTC}
\end{subfigure}
\begin{subfigure}[t]{0.1\textwidth}
\centering
\subcaption*{$40.071$ dB}
\includegraphics[scale=0.3]{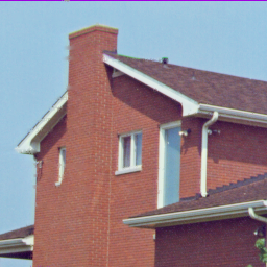}
\subcaption*{STTC}
\end{subfigure}
\begin{subfigure}[t]{0.1\textwidth}
\centering
\subcaption*{$38.291$ dB}
\includegraphics[scale=0.3]{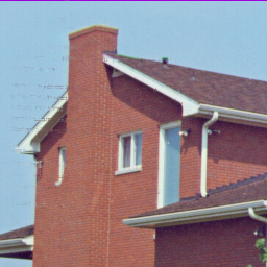}
\subcaption*{TRNNM}
\end{subfigure}
\caption{The recovery results of \emph{house}, where the missing position is a text-like mask.}
\label{result-image2}
\end{figure}

\begin{figure}[htbp]
\centering
\begin{subfigure}[t]{0.1\textwidth}
\centering
\subcaption*{Original}
\includegraphics[scale=0.2]{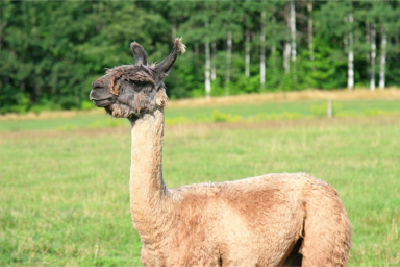}
\end{subfigure}
\begin{subfigure}[t]{0.1\textwidth}
\centering
\subcaption*{Observed}
\includegraphics[scale=0.2]{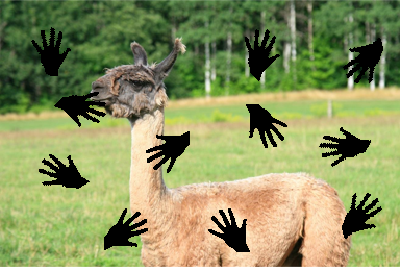}
\end{subfigure}

\begin{subfigure}[t]{0.1\textwidth}
\centering
\subcaption*{$34.061$ dB}
\includegraphics[scale=0.2]{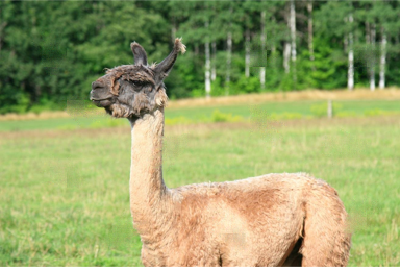}
\end{subfigure}
\begin{subfigure}[t]{0.1\textwidth}
\centering
\subcaption*{$24.980$ dB}
\includegraphics[scale=0.2]{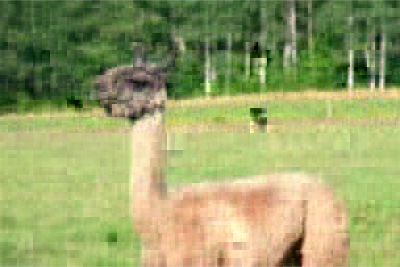}
\end{subfigure}
\begin{subfigure}[t]{0.1\textwidth}
\centering
\subcaption*{$32.404$ dB}
\includegraphics[scale=0.2]{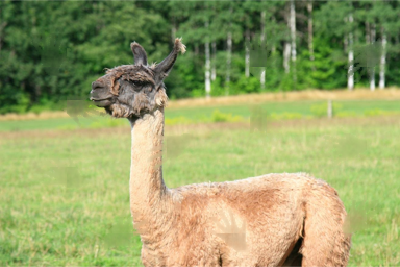}
\end{subfigure}
\begin{subfigure}[t]{0.1\textwidth}
\centering
\subcaption*{$33.025$ dB}
\includegraphics[scale=0.2]{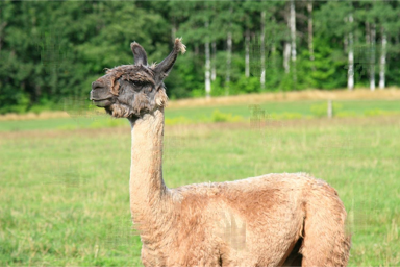}
\end{subfigure}
\begin{subfigure}[t]{0.1\textwidth}
\centering
\subcaption*{$28.969$ dB}
\includegraphics[scale=0.2]{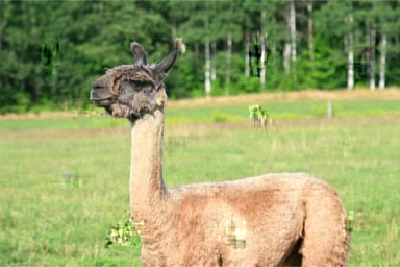}
\end{subfigure}
\begin{subfigure}[t]{0.1\textwidth}
\centering
\subcaption*{$33.152$ dB}
\includegraphics[scale=0.2]{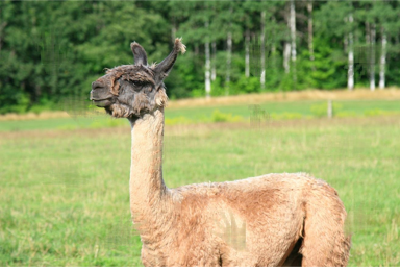}
\end{subfigure}
\begin{subfigure}[t]{0.1\textwidth}
\centering
\subcaption*{$32.731$ dB}
\includegraphics[scale=0.2]{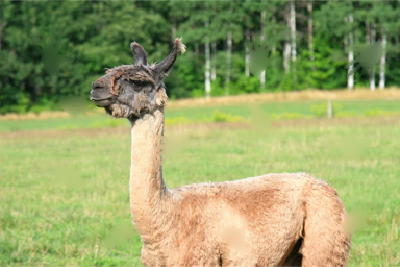}
\end{subfigure}
\begin{subfigure}[t]{0.1\textwidth}
\centering
\subcaption*{$32.866$ dB}
\includegraphics[scale=0.2]{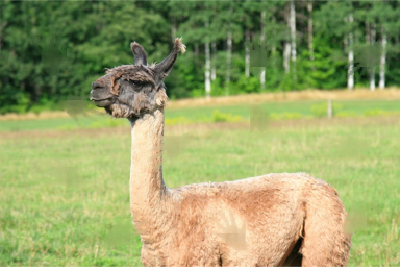}
\end{subfigure}

\begin{subfigure}[t]{0.1\textwidth}
\centering
\subcaption*{$15.268$ dB}
\includegraphics[scale=0.2]{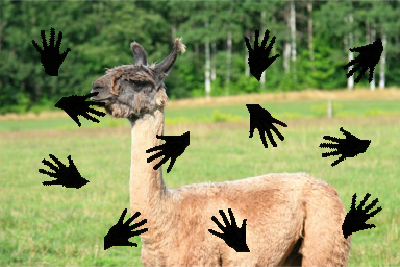}
\subcaption*{TRBU}
\end{subfigure}
\begin{subfigure}[t]{0.1\textwidth}
\centering
\subcaption*{$22.554$ dB}
\includegraphics[scale=0.2]{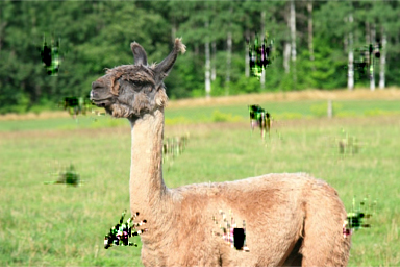}
\subcaption*{TR-ALS}
\end{subfigure}
\begin{subfigure}[t]{0.1\textwidth}
\centering
\subcaption*{$33.045$ dB}
\includegraphics[scale=0.2]{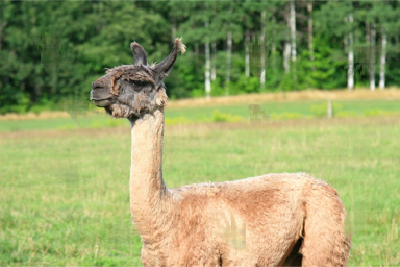}
\subcaption*{SiLRTC-TT}
\end{subfigure}
\begin{subfigure}[t]{0.1\textwidth}
\centering
\subcaption*{$33.025$ dB}
\includegraphics[scale=0.2]{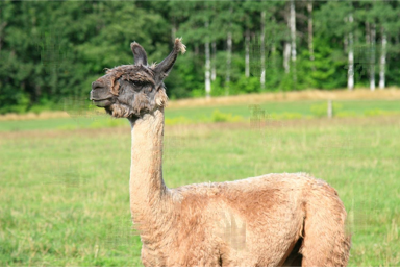}
\subcaption*{LRTC-TNN}
\end{subfigure}
\begin{subfigure}[t]{0.1\textwidth}
\centering
\subcaption*{$28.969$ dB}
\includegraphics[scale=0.2]{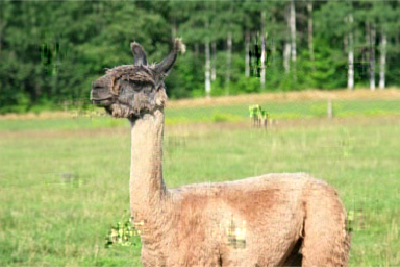}
\subcaption*{FBCP}
\end{subfigure}
\begin{subfigure}[t]{0.1\textwidth}
\centering
\subcaption*{$33.152$ dB}
\includegraphics[scale=0.2]{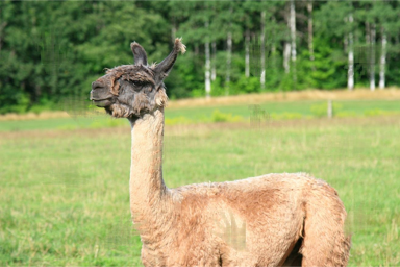}
\subcaption*{HaLRTC}
\end{subfigure}
\begin{subfigure}[t]{0.1\textwidth}
\centering
\subcaption*{$32.731$ dB}
\includegraphics[scale=0.2]{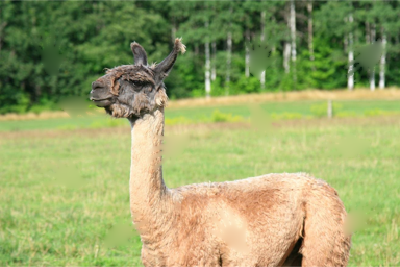}
\subcaption*{STTC}
\end{subfigure}
\begin{subfigure}[t]{0.1\textwidth}
\centering
\subcaption*{$32.971$ dB}
\includegraphics[scale=0.2]{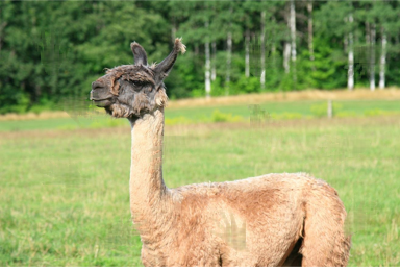}
\subcaption*{TRNNM}
\end{subfigure}
\caption{The recovery results of \emph{llama}, where the missing positions are a few palm-shaped graffiti.}
\label{result-image3}
\end{figure}

The image recovery results for \emph{house} and \emph{llama} are shown in Fig. \ref{result-image2} and Fig. \ref{result-image3}, respectively. The middle row shows the the recovery results of high-order tensor completions with VDT and the bottom row shows the recovery from directly completing the original images. Besides, it is apparent from Fig. \ref{result-image2} and Fig. \ref{result-image3} that TRBU is more capable of recovering high-order tensors.

\subsection{Real-world Videos}

In this section we use two videos to test the algorithms and perform $5$ experiments for each video. The first video called \emph{explosion} is an explosion shot by a high speed camera \footnote{http://www.newcger.com/shipinsucai/5786.html}. We selected its $81$st to $180$th frames and downsampled each frame to size $80\times 120\times 3$. It is further converted into a $9$-order tensor of size $8\times8\times4\times4\times10\times3\times4\times5\times5$ by VDT operation.

The second one is a color video that describes the activity of a bunch of chickens \footnote{https://pixabay.com/videos/id-10685/}. We downsampled each frame to size $72\times128\times3$ and finally get a tensor of size $8\times8\times4\times6\times6\times3\times4\times5\times5$ by VDT manipulation. The TR rank for TR-ALS is $12$ due to machine memory limit. The maximal CP rank for FBCP is limited by $40$. The sampling rate of two videos is $10\%$. We set $\beta=1.05$ and $\mu_0=1\times10^{-3.7}$ in this experiment.

\begin{figure}[htbp]
\centering
\begin{subfigure}[t]{0.11\textwidth}
\centering
\subcaption*{Original}
\includegraphics[scale=0.6]{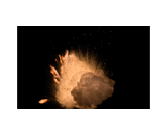}
\end{subfigure}
\begin{subfigure}[t]{0.11\textwidth}
\centering
\subcaption*{Observed}
\includegraphics[scale=0.6]{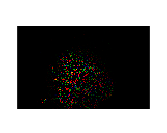}
\end{subfigure}

\begin{subfigure}[t]{0.11\textwidth}
\centering
\subcaption*{$37.256$ dB}
\includegraphics[scale=0.7]{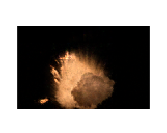}
\end{subfigure}
\begin{subfigure}[t]{0.11\textwidth}
\centering
\subcaption*{$32.281$ dB}
\includegraphics[scale=0.7]{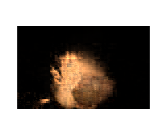}
\end{subfigure}
\begin{subfigure}[t]{0.11\textwidth}
\centering
\subcaption*{$19.759$ dB}
\includegraphics[scale=0.7]{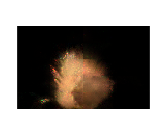}
\end{subfigure}
\begin{subfigure}[t]{0.11\textwidth}
\centering
\subcaption*{$17.870$ dB}
\includegraphics[scale=0.7]{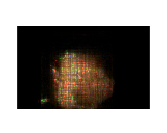}
\end{subfigure}
\begin{subfigure}[t]{0.11\textwidth}
\centering
\subcaption*{$32.665$ dB}
\includegraphics[scale=0.7]{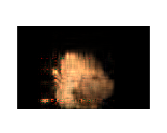}
\end{subfigure}
\begin{subfigure}[t]{0.11\textwidth}
\centering
\subcaption*{$24.200$ dB}
\includegraphics[scale=0.7]{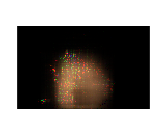}
\end{subfigure}
\begin{subfigure}[t]{0.11\textwidth}
\centering
\subcaption*{$24.921$ dB}
\includegraphics[scale=0.7]{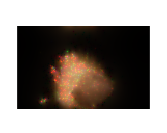}
\end{subfigure}
\begin{subfigure}[t]{0.11\textwidth}
\centering
\subcaption*{$15.064$ dB}
\includegraphics[scale=0.7]{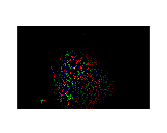}
\end{subfigure}

\begin{subfigure}[t]{0.11\textwidth}
\centering
\subcaption*{$21.023$ dB}
\includegraphics[scale=0.7]{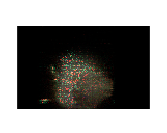}
\subcaption*{TRBU}
\end{subfigure}
\begin{subfigure}[t]{0.11\textwidth}
\centering
\subcaption*{$22.295$ dB}
\includegraphics[scale=0.7]{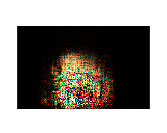}
\subcaption*{TR-ALS}
\end{subfigure}
\begin{subfigure}[t]{0.11\textwidth}
\centering
\subcaption*{$31.769$ dB}
\includegraphics[scale=0.7]{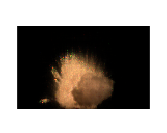}
\subcaption*{SiLRTC-TT}
\end{subfigure}
\begin{subfigure}[t]{0.11\textwidth}
\centering
\subcaption*{$31.042$ dB}
\includegraphics[scale=0.7]{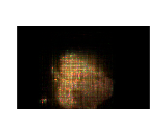}
\subcaption*{LRTC-TNN}
\end{subfigure}
\begin{subfigure}[t]{0.11\textwidth}
\centering
\subcaption*{$32.804$ dB}
\includegraphics[scale=0.7]{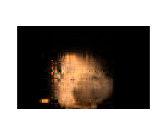}
\subcaption*{FBCP}
\end{subfigure}
\begin{subfigure}[t]{0.11\textwidth}
\centering
\subcaption*{$24.018$ dB}
\includegraphics[scale=0.7]{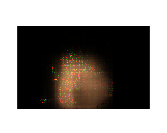}
\subcaption*{HaLRTC}
\end{subfigure}
\begin{subfigure}[t]{0.11\textwidth}
\centering
\subcaption*{$30.605$ dB}
\includegraphics[scale=0.7]{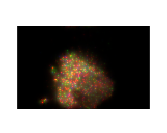}
\subcaption*{STTC}
\end{subfigure}
\begin{subfigure}[t]{0.11\textwidth}
\centering
\subcaption*{$15.594$ dB}
\includegraphics[scale=0.7]{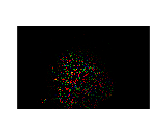}
\subcaption*{TRNNM}
\end{subfigure}
\caption{The recovery results of the last frame of \emph{explosion}, where the evaluation is based on the whole video.}
\label{result-video1}
\end{figure}

\begin{figure}[htbp]
\centering
\begin{subfigure}[t]{0.11\textwidth}
\centering
\subcaption*{Original}
\includegraphics[scale=0.7]{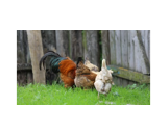}
\end{subfigure}
\begin{subfigure}[t]{0.11\textwidth}
\centering
\subcaption*{Observed}
\includegraphics[scale=0.7]{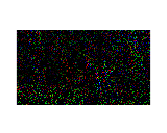}
\end{subfigure}

\begin{subfigure}[t]{0.11\textwidth}
\centering
\subcaption*{$29.337$ dB}
\includegraphics[scale=0.7]{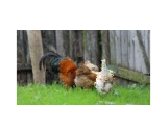}
\end{subfigure}
\begin{subfigure}[t]{0.11\textwidth}
\centering
\subcaption*{$23.527$ dB}
\includegraphics[scale=0.7]{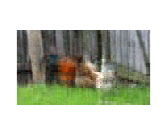}
\end{subfigure}
\begin{subfigure}[t]{0.11\textwidth}
\centering
\subcaption*{$10.736$ dB}
\includegraphics[scale=0.7]{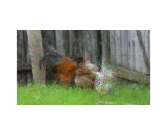}
\end{subfigure}
\begin{subfigure}[t]{0.11\textwidth}
\centering
\subcaption*{$12.148$ dB}
\includegraphics[scale=0.7]{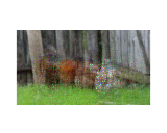}
\end{subfigure}
\begin{subfigure}[t]{0.11\textwidth}
\centering
\subcaption*{$22.849$ dB}
\includegraphics[scale=0.7]{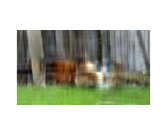}
\end{subfigure}
\begin{subfigure}[t]{0.11\textwidth}
\centering
\subcaption*{$19.570$ dB}
\includegraphics[scale=0.7]{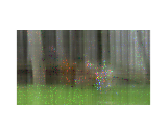}
\end{subfigure}
\begin{subfigure}[t]{0.11\textwidth}
\centering
\subcaption*{$23.690$ dB}
\includegraphics[scale=0.7]{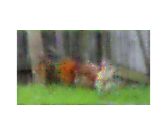}
\end{subfigure}
\begin{subfigure}[t]{0.11\textwidth}
\centering
\subcaption*{$25.918$ dB}
\includegraphics[scale=0.7]{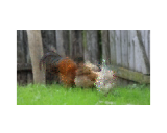}
\end{subfigure}

\begin{subfigure}[t]{0.11\textwidth}
\centering
\subcaption*{$11.672$ dB}
\includegraphics[scale=0.7]{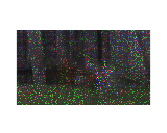}
\subcaption*{TRBU}
\end{subfigure}
\begin{subfigure}[t]{0.11\textwidth}
\centering
\subcaption*{$23.573$ dB}
\includegraphics[scale=0.7]{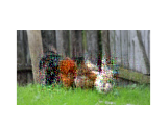}
\subcaption*{TR-ALS}
\end{subfigure}
\begin{subfigure}[t]{0.11\textwidth}
\centering
\subcaption*{$27.047$ dB}
\includegraphics[scale=0.7]{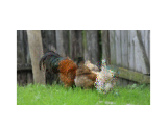}
\subcaption*{SiLRTC-TT}
\end{subfigure}
\begin{subfigure}[t]{0.11\textwidth}
\centering
\subcaption*{$23.749$ dB}
\includegraphics[scale=0.7]{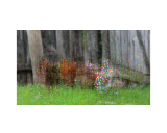}
\subcaption*{LRTC-TNN}
\end{subfigure}
\begin{subfigure}[t]{0.11\textwidth}
\centering
\subcaption*{$22.849$ dB}
\includegraphics[scale=0.7]{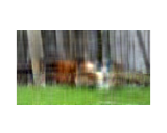}
\subcaption*{FBCP}
\end{subfigure}
\begin{subfigure}[t]{0.11\textwidth}
\centering
\subcaption*{$19.568$ dB}
\includegraphics[scale=0.7]{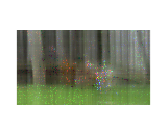}
\subcaption*{HaLRTC}
\end{subfigure}
\begin{subfigure}[t]{0.11\textwidth}
\centering
\subcaption*{$23.601$ dB}
\includegraphics[scale=0.7]{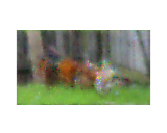}
\subcaption*{STTC}
\end{subfigure}
\begin{subfigure}[t]{0.11\textwidth}
\centering
\subcaption*{$7.424$ dB}
\includegraphics[scale=0.7]{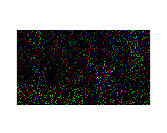}
\subcaption*{TRNNM}
\end{subfigure}
\caption{The recovery results of the first frame of \emph{cock}, where the evaluation is based on the whole video.}
\label{result-video2}
\end{figure}

We conducted the experiments for two whole videos. For each video, we show the recovery result of one frame in Fig. \ref{result-video1} and Fig. \ref{result-video2}. The middle row shows the recovery results for high-order tensor completion using VDT, and the bottom row shows the recovery results, in which case the original images are directly completed. The limited CP rank may deteriorate the performances of FBCP. This also implies huge storage requirement of FBCP. The TR-ALS is unable to effectively handle large scale data since it costs too much time. The TRBU has much better recovery quality among all methods and is efficient at large scale data completion.

\section{Conclusion}
\label{section-conclusion}

We study the tensor ring decomposition and propose a weighted sum of nuclear norm minimization model for tensor ring completion. Meanwhile, a recovery guarantee for TR completion under random sampling scheme is provided and proved. To verify the effectiveness of the proposed model, a method based on ADMM, namely TRBU, is proposed to tackle this problem. The results of experiments on synthetic data not only verify the proposed sampling condition for TR completion but also show that the sampling bound is conservative and can be improved, which will become our future work. Experiments on real-world data further demonstrate the efficiency of TRBU over other state-of-art algorithms, especially for higher-order tensor completion.

 \appendix
\section{Proof of Lemma \ref{lemma-TRsip}}
\label{appendix1}

\begin{proof*}
We first recall a concentration-of-measure inequality that is important for our analysis.
\begin{lemma}[McDiarmid inequality \cite{warnke2016method}]
\label{lemma-mcd}
Let $\left\{X_1,\dotsc,X_n\right\}$ be independent random variables such that there are $a_i \leq X_i \leq b_i$, $c_i:=b_i-a_i$ and $c_i\leq C$, $\forall i=1,\dotsc,n$. Let $S$ be an arbitrary (implicit) function of the variables, e.g., the sum function, then for any $t>0$ there is
\begin{equation}
\label{McDineq}
\mathrm{P}\left(|S-\mathrm{E}\left(S\right)|>t\right)<2e^{-\frac{2t^2}{nC^2}},
\end{equation}
as long as this function changes in a bounded way, i.e., if $X_i$ is changed, the value of this function changes by at most $C$.
\end{lemma}

We consider the $i$-th TR factor $\mathcal{G}^{\left(i\right)}$. According to the identity $\sum^{n_i}_{j_i=1}H^2_{j_i}=1$, there is $\mathrm{E}\left(H^2_{j_i}\right)=1/n_i$, supposing that $H_{j_i}=\mathcal{G}^{\left(i\right)}_{tj_is}$. Let $X_{ts}=\mathcal{G}^{\left(i\right)}_{tj_is} \mathcal{G}^{\left(i\right)}_{tj'_is}$ and $S=\sum_{t}\sum_{s}X_{ts}$, obviously $\mathrm{E}\left(S\right)=0$ if $j_i\neq j'_i$, and we have $\mathrm{E}\left(S\right)=\sum^{r_i}_{t=1}\sum^{r_{i+1}}_{s=1}\mathrm{E}\left(H^2_{j_i}\right)=r_ir_{i+1}/n_i$ if $j_i= j'_i$.

The proof is as follows. From the above deductions it is clear that $\mathrm{E}\left(S\right)=\frac{r_ir_{i+1}}{n_i}1_{j_i=j'_i}$. According to the union bound $\max\left(\mathcal{G}^{\left(i\right)}\right)\leq \sqrt{\frac{\mu_{B_i}}{n_i}}$, the bound of $X_{ts}$ is $C=2\frac{\mu_{B_i}}{n_i}$. Incorporating Lemma \ref{lemma-mcd} we have $\mathrm{Pr}\left(|\langle \mathcal{G}^{\left(i\right)}_{:j_i:}\;,\mathcal{G}^{\left(i\right)}_{:j'_i:} \rangle -\frac{r_ir_{i+1}}{n_i}1_{j_i=j'_i}|> \frac{\lambda \mu_{B_i}\sqrt{r_ir_{i+1}}}{n_i}\right)<2e^{-\frac{\lambda^2}{2}}$, and let $\lambda$ be a proportion of $\sqrt{\ln\left(n_i\right)}$ we prove (\ref{4.2}) with probability at least $1-n_i^{-3}$ (say). Additionally, there is $\mu_i=O\left(\mu_{B_i}\sqrt{\ln\left(n_i\right)}\right)$.

Note that the above result is only for one core of TR, the total probability is $\prod^{d}_{i=1}\left(1-n_i^{-3}\right)$.
$\hfill \blacksquare$
\end{proof*}

\section{Proof of Lemma \ref{lemma-TRunfoldingsip}}
\label{appendix2}

\begin{proof*}
We use the first formulation of TR decomposition, i.e., $x_{j_1\dotsm j_d}=\operatorname{tr}\left(\boldsymbol{\Sigma}^{\left(1\right)}\mathbf{G}^{\left(1\right)}_{j_1}\dotsm\ \boldsymbol{\Sigma}^{\left(d\right)}\mathbf{G}^{\left(d\right)}_{j_d}\right)$. Note that every mode-$2$ slice of $\mathcal{G}^{\left(i\right)}$ has the same status when interacting with $\boldsymbol{\Sigma}_i$ and $\boldsymbol{\Sigma}_{i+1}$, then a substitution for the representation of TR factors is $\widetilde{\mathcal{G}}^{\left(i\right)}=\mathcal{W}^{\left(i\right)}\circledast\mathcal{G}^{\left(i\right)}$, where $\mathcal{W}^{\left(i\right)}_{:j_i:}=\operatorname{diag}\left(\sqrt{\boldsymbol{\Sigma}_i}\right)\operatorname{diag}\left(\sqrt{\boldsymbol{\Sigma}_{i+1}}\right)^{\mathrm{T}}$ holds for all mode-$2$ slices of $\mathcal{W}^{\left(i\right)}$. We use matrix $\mathbf{W}^{\left(i\right)}$ to denote any mode-$2$ slice of $\mathcal{W}^{\left(i\right)}$ for convenience.

We consider a $4$-order tensor and calculate the SVD for its TR unfolding. For simplicity, we denote by $\mathbf{U}$ the $s_1$-th mode-$1$ slice of $\widetilde{\mathcal{G}}^{\left(1\right)}$ and $\mathbf{V}$ the $t_3$-th mode-$3$ slice of $\widetilde{\mathcal{G}}^{\left(2\right)}$. Consequently, the $\ell_2$-norm of the mode-$2$ fiber of $\widetilde{\mathcal{G}}^{\left(1\right)}\overline{\otimes} \widetilde{\mathcal{G}}^{\left(2\right)}$ is
\begin{align*}
\sqrt{\sum_{i}\sum_{j}\left(\mathbf{U}\mathbf{V}^{\mathrm{T}}\right)^2_{ij}}=& \sqrt{\sum_{k}w^{\left(1\right)}_{s_1k}w^{\left(2\right)}_{kt_3}\sum_{k'}w^{\left(1\right)}_{s_1k'}w^{\left(2\right)}_{k't_3}\sum_{i}u_{ik}u_{ik'}\sum_{j}v_{jk}v_{jk'}} \\
=& \left(\mathbf{W}^{\left(1\right)}\mathbf{W}^{\left(2\right)}\right)_{s_1t_3}
\end{align*}
by using the orthonormal condition of $\mathbf{U}$ and $\mathbf{V}$. Thus the $\ell_2$-norms of mode-$2$ fibers of $\widetilde{\mathcal{G}}^{\left(1\right)}\overline{\otimes} \widetilde{\mathcal{G}}^{\left(2\right)}$ are $\mathbf{W}^{\left(1\right)}\mathbf{W}^{\left(2\right)}=\operatorname{tr}\left(\boldsymbol{\Sigma}_2\right)\operatorname{diag}\left(\sqrt{\boldsymbol{\Sigma}_1}\right)\operatorname{diag}\left(\sqrt{\boldsymbol{\Sigma}_3}\right)^{\mathrm{T}}$ and the representation of the TR unfolding is
\begin{align*}
\mathbf{X}_{\left\{1,2\right\}}=\left[\left(\mathcal{W}^{\left(1\right)}\overline{*}\mathcal{W}^{\left(2\right)}\right)\circledast\mathcal{U}^{\left\{1,2\right\}}\right]_{\left(2\right)'}\left[\left(\mathcal{W}^{\left(3\right)}\overline{*}\mathcal{W}^{\left(4\right)}\right)\circledast\mathcal{V}^{\left\{3,2\right\}}\right]_{\left(2\right)}^{\mathrm{T}},
\end{align*}
where $\overline{*}$ is the slice-wise matrix product acting on corresponding mode-$2$ slices, operators $\left(\cdot\right)_{\left(2\right)'}$ and $\left(\cdot\right)_{\left(2\right)}$ unfold a TR factor into a matrix with permuted order $\left[2,3,1\right]^{\mathrm{T}}$ and $\left[2,1,3\right]^{\mathrm{T}}$, respectively. We derive $\mathbf{X}_{\left\{1,2\right\}}=\mathcal{U}^{\left\{1,2\right\}}_{\left(2\right)'}\boldsymbol{\Sigma}_{13}\mathcal{V}^{{\left\{3,2\right\}}^{\mathrm{T}}}_{\left(2\right)}$, where $\mathcal{U}^{\left\{1,2\right\}}=\mathcal{G}^{\left(1\right)}\overline{\otimes} \mathcal{G}^{\left(2\right)}$, $\mathcal{V}^{\left\{3,2\right\}}=\mathcal{G}^{\left(3\right)}\overline{\otimes} \mathcal{G}^{\left(4\right)}$ and
\begin{equation*}
\begin{split}
\boldsymbol{\Sigma}_{13}=& \operatorname{diag}\left(\overrightarrow{\left(\mathbf{W}^{\left(1\right)}\mathbf{W}^{\left(2\right)}\right)}\right)\circledast \operatorname{diag}\left(\downarrow\left(\mathbf{W}^{\left(3\right)}\mathbf{W}^{\left(4\right)}\right)\right) \\
=& \operatorname{tr}\left(\boldsymbol{\Sigma}_2\right)\operatorname{diag}\left(\overrightarrow{\left(\operatorname{diag}\left(\sqrt{\boldsymbol{\Sigma}_1}\right)\operatorname{diag}\left(\sqrt{\boldsymbol{\Sigma}_3}\right)^{\mathrm{T}}\right)}\right)\circledast \\& \operatorname{tr}\left(\boldsymbol{\Sigma}_4\right)\operatorname{diag}\left(\downarrow\left(\operatorname{diag}\left(\sqrt{\boldsymbol{\Sigma}_3}\right)\operatorname{diag}\left(\sqrt{\boldsymbol{\Sigma}_1}\right)^{\mathrm{T}}\right)\right) \\
=& \operatorname{tr}\left(\boldsymbol{\Sigma}_2\right)\operatorname{tr}\left(\boldsymbol{\Sigma}_4\right)\operatorname{diag}\left(\downarrow\left(\sqrt{\boldsymbol{\sigma}_3}\sqrt{\boldsymbol{\sigma}_1}^{\mathrm{T}}\right)\right)\circledast \\
& \operatorname{diag}\left(\downarrow\left(\sqrt{\boldsymbol{\sigma}_3}\sqrt{\boldsymbol{\sigma}_1}^{\mathrm{T}}\right)\right) \\
=& \operatorname{tr}\left(\boldsymbol{\Sigma}_2\right)\operatorname{tr}\left(\boldsymbol{\Sigma}_4\right)\operatorname{diag}\left(\left(\sqrt{\boldsymbol{\sigma}_1}\otimes \sqrt{\boldsymbol{\sigma}_3}\right)\circledast \left(\sqrt{\boldsymbol{\sigma}_1}\otimes \sqrt{\boldsymbol{\sigma}_3}\right)\right) \\
=& \operatorname{tr}\left(\boldsymbol{\Sigma}_2\right)\operatorname{tr}\left(\boldsymbol{\Sigma}_4\right)\operatorname{diag}\left(\boldsymbol{\sigma}_1\otimes \boldsymbol{\sigma}_3\right) \\
=& \operatorname{tr}\left(\boldsymbol{\Sigma}_2\right)\operatorname{tr}\left(\boldsymbol{\Sigma}_4\right)\boldsymbol{\Sigma}_1\otimes \boldsymbol{\Sigma}_3.
\end{split}
\end{equation*}
Here the operator $\downarrow{\left(\cdot\right)}:=\operatorname{vec}\left(\cdot\right)$ rearranges a matrix into a vector column by column and $\overrightarrow{\left(\cdot\right)}:=\operatorname{vec}\left(\cdot^{\mathrm{T}}\right)$ rearranges a matrix into a vector row by row. To determine the rank of $\mathbf{X}_{\left\{1,2\right\}}$, we have $\operatorname{rank}\left(\boldsymbol{\Sigma}_{13}\right)=\operatorname{rank}\left(\boldsymbol{\Sigma}_1\right)\operatorname{rank}\left(\boldsymbol{\Sigma}_3\right)=r_1r_3$.

The next step is to verify the orthogonality of $\mathcal{U}^{\left\{1,2\right\}}$ and $\mathcal{V}^{\left\{3,2\right\}}$. Since
\begin{align*}
\sum_{k}w^{\left(1\right)}_{r_1k}w^{\left(2\right)}_{kr_3}\sum_{t}w^{\left(1\right)}_{r_1t}w^{\left(2\right)}_{tr_3}\sum_{i}u'_{ik}u''_{it}\sum_{j}v'_{jk}v''_{jt}\equiv 0,
\end{align*}
where $\mathbf{U}'\neq \mathbf{U}''$ or $\mathbf{V}'\neq \mathbf{V}''$, which means two pairs of slices are not allowed to be the same at the same time. With this expression it is clear that both $\mathcal{U}^{\left\{1,2\right\}}=\mathcal{G}^{\left(1\right)}\overline{\otimes} \mathcal{G}^{\left(2\right)}$ and $\mathcal{V}^{\left\{3,2\right\}}=\mathcal{G}^{\left(3\right)}\overline{\otimes} \mathcal{G}^{\left(4\right)}$ are orthogonal.

Generally there are $\mathbf{X}_{\left\{k,l\right\}}=\mathcal{U}^{\left\{k,l\right\}}_{\left(2\right)'}\boldsymbol{\Sigma}_{k,k+l}\mathcal{V}^{{\left\{k+l,d-l\right\}}^{\mathrm{T}}}_{\left(2\right)}$, where $\mathcal{U}^{\left\{k,l\right\}}= \overline{\otimes}^{k+l-1}_{i=k}\mathcal{G}^{\left(i\right)}$, $\mathcal{V}^{\left\{k+l,d-l\right\}}= \overline{\otimes}^{k-1}_{i=k+l}\mathcal{G}^{\left(i\right)}$ and $\boldsymbol{\Sigma}_{k,k+l}= \prod_{i\neq k,\;k+l}\operatorname{tr}\left(\boldsymbol{\Sigma}_i\right)\boldsymbol{\Sigma}_k\otimes \boldsymbol{\Sigma}_{k+l}$. The rank is given by $\operatorname{rank}\left(\mathbf{X}_{\left\{k,l\right\}}\right)=r_kr_{k+l}$.

To calculate the $\ell_2$-norm of the mode-$2$ fiber of $\overline{\otimes}^{n+l-1}_{k=n}\mathcal{G}^{\left(k\right)}$, we consider a simple case in which two factors are contracted. The $\left(r_1,r_3\right)$-th mode-$2$ fiber of $\mathcal{G}^{\left(1\right)}\overline{\otimes} \mathcal{G}^{\left(2\right)}$ can be written as $\left(\mathcal{G}^{\left(1\right)}\overline{\otimes} \mathcal{G}^{\left(2\right)}\right)_{r_1:r_3} =\downarrow{\left(\mathcal{G}^{\left(1\right)}_{r_1::}\mathcal{G}^{\left(2\right)}_{::r_3}\right)}$, and hence the $\ell_2$-norm of the fiber is equal to the $\mathrm{F}$-norm of the matrix which is $\lVert \mathcal{G}^{\left(1\right)}_{r_1::}\mathcal{G}^{\left(2\right)}_{::r_3} \rVert_{\mathrm{F}}=\lVert \mathbf{E}_{r_2} \rVert_{\mathrm{F}}=r_2$. This equation is because mode-$2$ fibers of $\mathcal{G}^{\left(1\right)}$ and $\mathcal{G}^{\left(2\right)}$ are orthonormal. Let $\mathcal{G}^{\left(1\right)}\overline{\otimes} \mathcal{G}^{\left(2\right)}$ be a contracted factor and recursively repeat the above procedure, we prove that the $\ell_2$-norm of the mode-$2$ fiber of $\overline{\otimes}^{k+l-1}_{i=k}\mathcal{G}^{\left(i\right)}$ is $\prod^{k+l-1}_{i=k+1}r_i$. Therefore, the left and right singular matrices of $\mathbf{X}_{\left\{k,l\right\}}$ are $\mathcal{U}^{\left\{k,l\right\}}=\prod^{k+l-1}_{i=k+1}\frac{1}{r_i}\overline{\otimes}^{k+l-1}_{i=k}\mathcal{G}^{\left(i\right)}$ and $\mathcal{V}^{\left\{k+l,d-l\right\}}=\prod^{k-1}_{i=k+l+1}\frac{1}{r_i}\overline{\otimes}^{k-1}_{i=k+l}\mathcal{G}^{\left(i\right)}$.

Subsequently, we calculate the variable expectation. Similar to the proof of Definition \ref{lemma-TRsip}, let $H_{s\bar{i}t}=\left({\overline{\otimes}}^{k+l-1}_{i=k}\mathcal{G}^{\left(i\right)}\right)_{s\bar{i}t}=\left(\prod^{k+l-1}_{i=k} \mathcal{G}^{\left(i\right)}_{:i_k:}\right)_{st}$, there is $\mathrm{E}\left(H^2_{s\bar{i}t}\right)=\prod^{k+l-1}_{i=k+1}r_i^2/\prod^{k+l-1}_{i=k}n_i$, where $\bar{i}\in \left\{1,\dotsc,1+\sum_{i=k}^{k+l-1}{\left( n_i-1 \right)\prod_{j=i}^{i-1}{n_j}}\right\}$. We define $S=\langle \mathcal{U}_{:\bar{i}:},\mathcal{U}_{:\bar{j}:} \rangle=\langle \left({\overline{\otimes}}^{k+l-1}_{i=k}\mathcal{G}^{\left(i\right)}\right)_{:\bar{i}:}, \left({\overline{\otimes}}^{k+l-1}_{i=k}\mathcal{G}^{\left(i\right)}\right)_{:\bar{j}:} \rangle=\sum_{s}\sum_{t}H_{s\bar{i}t}H_{s\bar{j}t}$, then $\mathrm{E}\left(S\right)=\left(r_kr_{k+l}\prod^{k+l-1}_{i=k}r_i^2\prod^{k+l-1}_{i=k}n_i^{-1}\right)1_{i=j}$, where the definition of $j$ is the same with that of $i$ above. Performing the normalization we have a modification $\mathrm{E}\left(S\right)=\left(r_kr_{k+l}\prod^{k+l-1}_{i=k}n_i^{-1}\right)1_{i=j}$.

Assuming that only two factors are allowed to be contracted at a time and we start the contraction from the $k$-th core. Notice the normalization, and the variable bound can be calculated by the following recursive formula $C_{\left\{k,l\right\}}=C_{\left\{k,l-1\right\}}\frac{\mu_{B_{k+l-1}}}{n_{k+l-1}}$, where $C_{\left\{k,1\right\}}=2\frac{\mu_{B_k}}{n_k}$. This implies $C_{\left\{k,l\right\}}=2\prod^{k+l-1}_{i=k}\frac{\mu_{B_i}}{n_i}$. It is trivial to verify that $C_{\left\{k,l\right\}}<C_{\left\{k,1\right\}}$ for enough large $\left\{n_i\right\}$.

According to Lemma \ref{lemma-mcd}, we have $\mathrm{Pr}\left(|\langle \mathcal{U}^{\left\{k,l\right\}}_{:\bar{i}:},\mathcal{U}^{\left\{k,l\right\}}_{:\bar{j}:} \rangle-\frac{r_kr_{k+l}}{\prod^{k+l-1}_{i=k}n_i}1_{\bar{i}=\bar{j}}|> t \right)<2e^{-\frac{2t^2}{r_kr_{k+l}C_{\left\{k,l\right\}}^2}}$. Let $t=\lambda \sqrt{r_kr_{k+l}}\prod^{k+l-1}_{i=k}\frac{\mu_{B_i}}{n_i}$, the right term becomes $2e^{-\frac{\lambda^2}{2}}$. Choosing $\lambda=O\left(\sqrt{\sum^{k+l-1}_{i=k}\ln n_i}\right)$, inequality $|\langle \mathcal{U}^{\left\{k,l\right\}}_{:\bar{i}:},\mathcal{U}^{\left\{k,l\right\}}_{:\bar{j}:} \rangle-\frac{r_kr_{k+l}}{\prod^{k+l-1}_{i=k}n_i}1_{\bar{i}=\bar{j}}|\leq \lambda \sqrt{r_kr_{k+l}}\prod^{k+l-1}_{i=k}\frac{\mu_{B_i}}{n_i}$ holds with probability at least $1-\prod^{k+l-1}_{i=k}n_i^{-3}$.

The proof of inequality about $\mathcal{V}^{\left\{k+l,d-l\right\}}$ is similar to that of $\mathcal{U}^{\left\{k,l\right\}}$ and hence is skipped.

Let $X_{st}=\left({\overline{\otimes}}^{k+l-1}_{i=k}\mathcal{G}^{\left(i\right)}\right)_{s\bar{i}t} \left({\overline{\otimes}}^{k-1}_{i=k+l}\mathcal{G}^{\left(i\right)}\right)_{s\bar{j}t}$, it is evident that $\mathrm{E}\left(X_{st}\right)\equiv 0$, and the variable bound satisfies $C=C_{\left\{k,l\right\}}C_{\left\{k+l,d-l\right\}}=2\prod^d_{i=1}\frac{\mu_{B_i}}{n_i}$. Plugging $t=\sqrt{r_kr_{k+l}}\prod^d_{i=1}\mu_{B_i}n_i^{-1/2}$ into $\mathrm{Pr}\left(|\langle \mathcal{U}^{\left\{k,l\right\}}_{:\bar{i}:},\mathcal{V}^{\left\{k+l,d-l\right\}}_{:\bar{j}:} \rangle|> t \right)<2e^{-\frac{2t^2}{r_kr_{k+l}C^2}}$, we prove $|\langle \mathcal{U}^{\left\{k,l\right\}}_{:\bar{i}:},\mathcal{V}^{\left\{k+l,d-l\right\}}_{:\bar{j}:} \rangle| \leq \frac{\sqrt{r_kr_{k+l}}\prod^d_{i=1}\mu_{B_i}}{\prod^d_{i=1}\sqrt{n_i}}$ holds with probability at least $1-e^{-\frac{1}{2}\prod^{d}_{i=1}n_i}$.

With the above deduction, the following inequalities
\begin{equation*}
\left\{
\begin{aligned}
& |\langle \mathcal{U}^{\left\{k,l\right\}}_{:\bar{i}:},\mathcal{U}^{\left\{k,l\right\}}_{:\bar{j}:} \rangle-\frac{r_kr_{k+l}}{\prod^{k+l-1}_{i=k}n_i}1_{\bar{i}=\bar{j}}|\leq \frac{\mu'_{1kl}\sqrt{r_kr_{k+l}}}{\prod^{k+l-1}_{i=k}n_i} \\
& |\langle \mathcal{V}^{\left\{k+l,d-l\right\}}_{:\bar{i}:},\mathcal{V}^{\left\{k+l,d-l\right\}}_{:\bar{j}:} \rangle-\frac{r_kr_{k+l}}{\prod^{k-1}_{i=k+l}n_i}1_{\bar{i}=\bar{j}}|\leq \frac{\mu'_{2kl}\sqrt{r_kr_{k+l}}}{\prod^{k-1}_{i=k+l}n_i}
\end{aligned}
\right.
\end{equation*}
hold with probabilities at least $1-\prod^{k+l-1}_{i=k}n_i^{-3}$ and $1-\prod^{k-1}_{i=k+l}n_i^{-3}$, respectively. Inequality $|\langle \mathcal{U}^{\left\{k,l\right\}}_{:\bar{i}:},\mathcal{V}^{\left\{k+l,d-l\right\}}_{:\bar{j}:} \rangle|\leq \frac{\mu''\sqrt{r_kr_{k+l}}}{\sqrt{\prod^{d}_{i=1}n_i}}$ holds with probability at least $1-e^{-\frac{1}{2}\prod^{d}_{i=1}n_i}$, where 
\begin{equation*}
\left\{
\begin{aligned}
\mu'_{1kl}=& O\left(\prod^{k+l-1}_{i=k}\mu_{B_i}\sqrt{\sum^{k+l-1}_{i=k}\ln n_i}\right),\;\mu'_{2kl}=O\left(\prod^{k-1}_{i=k+l}\mu_{B_i}\sqrt{\sum^{k-1}_{i=k+l}\ln n_i}\right) \\
\mu''=& \prod^{d}_{i=1}\mu_{B_i}
\end{aligned}
\right.
\end{equation*}
The proof is end.
$\hfill \blacksquare$
\end{proof*}

\section{Proof of Lemma \ref{lemma-dual certificate}}
\label{appendix3}

\begin{proof*}
Since Lemma \ref{lemma-TRunfoldingsip} indicates a TR unfolding has a unique SVD, we define the linear spaces $T_{il}=\{\mathbf{T}|\mathbf{T}=\prod^{i+l-1}_{k=i+1}\frac{1}{r_k}\mathcal{U}^{\left\{i,l\right\}}_{\left(2\right)'}\mathbf{Y}^{\mathrm{T}}+\prod^{i-1}_{k=i+l+1}\frac{1}{r_k}\mathbf{X}\mathcal{V}^{{\left\{i,l\right\}}^\mathrm{T}}_{\left(2\right)},\; \forall\;\mathbf{X},\mathbf{Y}\in\mathbb{R}^{\prod^{i+l-1}_{k=i}n_k\times\prod^{i-1}_{k=i+l}n_k}\}$ and $T_{il}^{\perp}$ as the orthogonal complement of $T_{il}$. The formulations of orthogonal projections $\mathscr{P}_{T_{il}}$ and $\mathscr{P}_{T_{il}^{\perp}}=\mathscr{I}-\mathscr{P}_{T_{il}}$ are $\mathscr{P}_{T_{il}}\left(\mathbf{X}\right)=\mathscr{P}_{U_{il}}\mathbf{X}+\mathbf{X}\mathscr{P}_{V_{il}}-\mathscr{P}_{U_{il}}\mathbf{X}\mathscr{P}_{V_{il}}$ and $\mathscr{P}_{T_{il}^{\perp}}\left(\mathbf{X}\right)=\left(\mathscr{I}-\mathscr{P}_{U_{il}}\right)\mathbf{X}\left(\mathscr{I}-\mathscr{P}_{V_{il}}\right)$, respectively.

To prove (\ref{condition1}), we verify it with the Rudelson selection estimate \cite{candes2010power} under the assumption of strong TR incoherence condition. Since $\lVert \mathscr{P}_{T_{il}}\mathscr{P}_{\Omega_{il}}\mathscr{P}_{T_{il}}-p\mathscr{P}_{T_{il}} \rVert_2\leq ap$ with probability at least $1-3\overline{n}_{il}^{-\beta}$ for any $\beta>1$ and $a=C_R\sqrt{\mu_0\overline{n}_{il}r_ir_{i+l}\beta\ln \left(\overline{n}_{il}\right)/m}<1$, note that $\mu_0\leq 1+\max\left\{\mu_{1il},\mu_{2il}\right\}/\sqrt{r_ir_{i+l}}\leq 1+\mu_{il}/\sqrt{r_ir_{i+l}}$. Applying this theorem with $a=1/2$ and $\beta=4$ gives the validation of (\ref{condition1}), where $m$ is required to be larger than $\max_{i} C\mu \overline{n}_{il}r_ir_{i+l}\ln\left(\overline{n}_{il}\right)$. The proof to (\ref{condition1}) is complete.

To prove the first condition in (\ref{condition2}), we construct the dual certificate via the Golf scheme introduced in \cite{candes2009exact}. Considering a union of $\Omega$ where $\Omega=\cup^{j_0}_{j=1}\Omega^j$ and $j_0=\lfloor \frac{5}{2}\log_2\left(\overline{n}_{kl}\right)+1 \rfloor$, and each $\Omega^j$ obeys the Bernoulli model $\Omega^j\sim \operatorname{Ber}\left(q_j\right)$ where $q_j=1-\left(1-p\right)^{1/j_0}$, $p=m\prod^{d}_{i=1}n_i^{-1}$. Inductively defining
\begin{equation*}
\left\{
\begin{aligned}
& \mathbf{Z}^0_{\left\{i,l\right\}}=\mathcal{U}^{\left\{i,l\right\}}_{\left(2\right)'}\mathcal{V}^{{\left\{i+l,d-l\right\}}^{\mathrm{T}}}_{\left(2\right)}=\mathscr{R}\left(\left\{\mathcal{G}\right\}\right)_{\left\{i,l\right\}} \\
& \mathbf{Y}^j_{\left\{i,l\right\}}=\sum^{j}_{k=1}q_k^{-1}\mathscr{P}_{\Omega^k_{il}}\mathscr{P}_{T_{il}}\left(\mathbf{Z}^{k-1}_{\left\{i,l\right\}}\right) \\
& \mathbf{Z}^j_{\left\{i,l\right\}}=\mathbf{Z}^0_{\left\{i,l\right\}}-\mathscr{P}_{T_{il}}\left(\mathbf{Y}^j_{\left\{i,l\right\}}\right)
\end{aligned}
\right.,\; i=1,\dotsc,\lceil d/2 \rceil,
\end{equation*}
which implies $\mathbf{Z}^j_{\left\{i,l\right\}}=\left(\mathscr{P}_{T_{il}}-q_j^{-1}\mathscr{P}_{T_{il}}\mathscr{P}_{\Omega^j_{il}}\mathscr{P}_{T_{il}}\right)\left(\mathbf{Z}^{j-1}_{\left\{i,l\right\}}\right)$ and $\mathscr{P}_{\Omega}\left(\mathcal{Y}\right)=\mathcal{Y}$, where $\mathscr{R}\left(\left\{\mathcal{G}\right\}\right)$ means contracting a TR whose singular value matrices are all identity matrices. Note that $\lVert \mathbf{Z}^{j}_{\left\{i,l\right\}}  \rVert_2\leq \lVert \mathscr{P}_{T_{il}}-q_{j}^{-1}\mathscr{P}_{T_{il}}\mathscr{P}_{\Omega^j_{il}}\mathscr{P}_{T_{il}} \rVert_2 \lVert \mathbf{Z}^{j-1}_{\left\{i,l\right\}} \rVert_2$, then
\begin{align*}
\lVert \mathscr{P}_{T_{il}}\left( \mathbf{Y}_{\left\{i,l\right\}} \right)-\mathcal{U}^{\left\{i,l\right\}}_{\left(2\right)'}\mathcal{V}^{{\left\{i+l,d-l\right\}}^{\mathrm{T}}}_{\left(2\right)} \rVert_2= \lVert \mathbf{Z}^{j_0}_{\left\{i,l\right\}}  \rVert_2\leq 2^{-j_0}\lVert \mathbf{Z}^{0}_{\left\{i,l\right\}} \rVert_{\mathrm{F}}\leq& 2^{-j_0}\sqrt{\underline{n}_{il}}\leq \frac{1}{2}\prod^{d}_{i=1}n_i^{-1}.
\end{align*}
The proof to the first condition of (\ref{condition2}) is complete. 

Then we prove the second condition of (\ref{condition2}). We deduce
\begin{align*}
\lVert \mathscr{P}_{T_{il}^{\perp}}\left(\mathbf{Y}^{j_0}_{\left\{i,l\right\}}\right) \rVert_2=& \lVert \sum^{j_0}_{j=1}q_j^{-1}\mathscr{P}_{T_{il}^{\perp}}\mathscr{P}_{\Omega^j_{il}}\mathscr{P}_{T_{il}}\left(\mathbf{Z}^{j-1}_{\left\{i,l\right\}}\right) \rVert_2 \\
\leq& \sum^{j_0}_{j=1}q_j^{-1}\lVert \mathscr{P}_{T_{il}^{\perp}}\mathscr{P}_{\Omega^j_{il}}\mathscr{P}_{T_{il}}\left(\mathbf{Z}^{j-1}_{\left\{i,l\right\}}\right) \rVert_2 \\
\leq& \sum^{j_0}_{j=1}\lVert \left(q_j^{-1}\mathscr{P}_{\Omega^j_{il}}-\mathscr{I}\right)\left(\mathbf{Z}^{j-1}_{\left\{i,l\right\}}\right) \rVert_2 \\
\leq& \sum^{j_0}_{j=1}2C_0\sqrt{q_j^{-1}\overline{n}_{il}\ln\left(\overline{n}_{il}\right)}\lVert \mathbf{Z}^{j-1}_{\left\{i,l\right\}} \rVert_{\ell_\infty} \\
\leq& \sum^{j_0}_{j=1}2^{2-j}C_0\sqrt{q_j^{-1}\overline{n}_{il}\ln\left(\overline{n}_{il}\right)}\lVert \mathbf{Z}^0_{\left\{i,l\right\}} \rVert_{\ell_\infty} \\
\leq& 4C_0\sqrt{\frac{\overline{n}_{il}\ln\left(\overline{n}_{il}\right)}{1-\left(1-p\right)^{1/j_0}}}\prod^{d}_{i=1}\frac{\mu_{B_i}r_i}{n_i} \\
\leq& 4C_0\sqrt{\frac{j_0\ln\left(\overline{n}_{il}\right)}{m\underline{n}_{il}}}\prod^{d}_{i=1}\mu_{B_i}r_i \\
\leq& 4\sqrt{3}C_0\frac{\log_2\left(\overline{n}_{il}\right)}{\sqrt{m\underline{n}_{il}}}\prod^{d}_{i=1}\mu_{B_i}r_i \\
\leq& \frac{1}{2}.
\end{align*}
The first inequality comes from the triangle inequality, the second inequality follows form $\mathscr{P}_{T_{il}^{\perp}}\mathscr{P}_{T_{il}}\equiv 0$, the third inequality is derived by bounding $\left(q_j^{-1}\mathscr{P}_{\Omega^j_{il}}-\mathscr{I}\right)\left(\mathbf{Z}^{j-1}_{\left\{i,l\right\}}\right) \rVert_2$ using Theorem 6.3 in \cite{candes2009exact}. The fifth inequality is a result of bounding $\lVert \mathbf{Z}^0_{\left\{i,l\right\}} \rVert_{\ell_\infty}$ which can be derived from the proof of Lemma \ref{lemma-TRunfoldingsip}. The last inequality requires $C_0$ and $m$ are larger enough. The proof to the second condition of (\ref{condition2}) is complete.
$\hfill \blacksquare$
\end{proof*}


\bibliographystyle{elsarticle-num} 
\bibliography{references_TRBU}

%\end{thebibliography}
\end{document}